\documentclass[10pt,twocolumn,letterpaper]{article}

\usepackage{cvpr}              % To produce the CAMERA-READY version
\usepackage[accsupp]{axessibility}

\usepackage[dvipsnames]{xcolor}

\usepackage{graphicx}
\usepackage{amsmath}
\usepackage{amssymb}
\usepackage{booktabs}

\usepackage{times}
\usepackage{epsfig}

\usepackage{verbatim}

\usepackage{algorithm}
\usepackage{algorithmicx}
\usepackage{algpseudocode}

\usepackage{color}

\usepackage{ctable}
\usepackage{makecell}
\usepackage{tabularx}
\usepackage{multirow}
\usepackage{multicol}
\usepackage{arydshln}

\usepackage{pifont}
\newcommand{\cmark}{\ding{51}}%
\newcommand{\xmark}{\ding{55}}%

\newcommand{\ourmethod}{DeClotH}
\definecolor{cvprblue}{rgb}{0.21,0.49,0.74}
\usepackage[pagebackref,breaklinks,colorlinks,citecolor=cvprblue]{hyperref}

\def\paperID{8156} % *** Enter the Paper ID here
\def\confName{CVPR}
\def\confYear{2025}

\title{\ourmethod: Decomposable 3D Cloth and Human Body Reconstruction \\ from a Single Image}

\author{
  Hyeongjin Nam$^{1,3}$ \hskip1.6em Donghwan Kim$^{1}$ \hskip1.6em Jeongtaek Oh$^{2}$ \hskip1.6em Kyoung Mu Lee$^{1,2}$ \vspace{+2mm} \\ 
   $^{1}$Dept. of ECE\&ASRI, $^{2}$IPAI, Seoul National University, $^{3}$KRAFTON   \\ 
   {\tt\small \{namhjsnu28, dh971106, ohjtgood, kyoungmu\}@snu.ac.kr} \\
   \small{\url{https://hygenie1228.github.io/DeClotH/}}
}

\begin{document}
\maketitle
\begin{abstract}
Most existing methods of 3D clothed human reconstruction from a single image treat the clothed human as a single object without distinguishing between cloth and human body.
In this regard, we present \textbf{\ourmethod}, which separately reconstructs 3D cloth and human body from a single image.
This task remains largely unexplored due to the extreme occlusion between cloth and the human body, making it challenging to infer accurate geometries and textures.
Moreover, while recent 3D human reconstruction methods have achieved impressive results using text-to-image diffusion models, directly applying such an approach to this problem often leads to incorrect guidance, particularly in reconstructing 3D cloth.
To address these challenges, we propose two core designs in our framework.  
First, to alleviate the occlusion issue, we leverage 3D template models of cloth and human body as regularizations, which provide strong geometric priors to prevent erroneous reconstruction by the occlusion.
Second, we introduce a cloth diffusion model specifically designed to provide contextual information about cloth appearance, thereby enhancing the reconstruction of 3D cloth.
Qualitative and quantitative experiments demonstrate that our proposed approach is highly effective in reconstructing both 3D cloth and the human body.
\end{abstract}

\section{Introduction}
\label{sec:introduction}
\begin{figure}[t!]
  \centering
  \vspace{+0.5em}
  \includegraphics[width=1.0\linewidth]{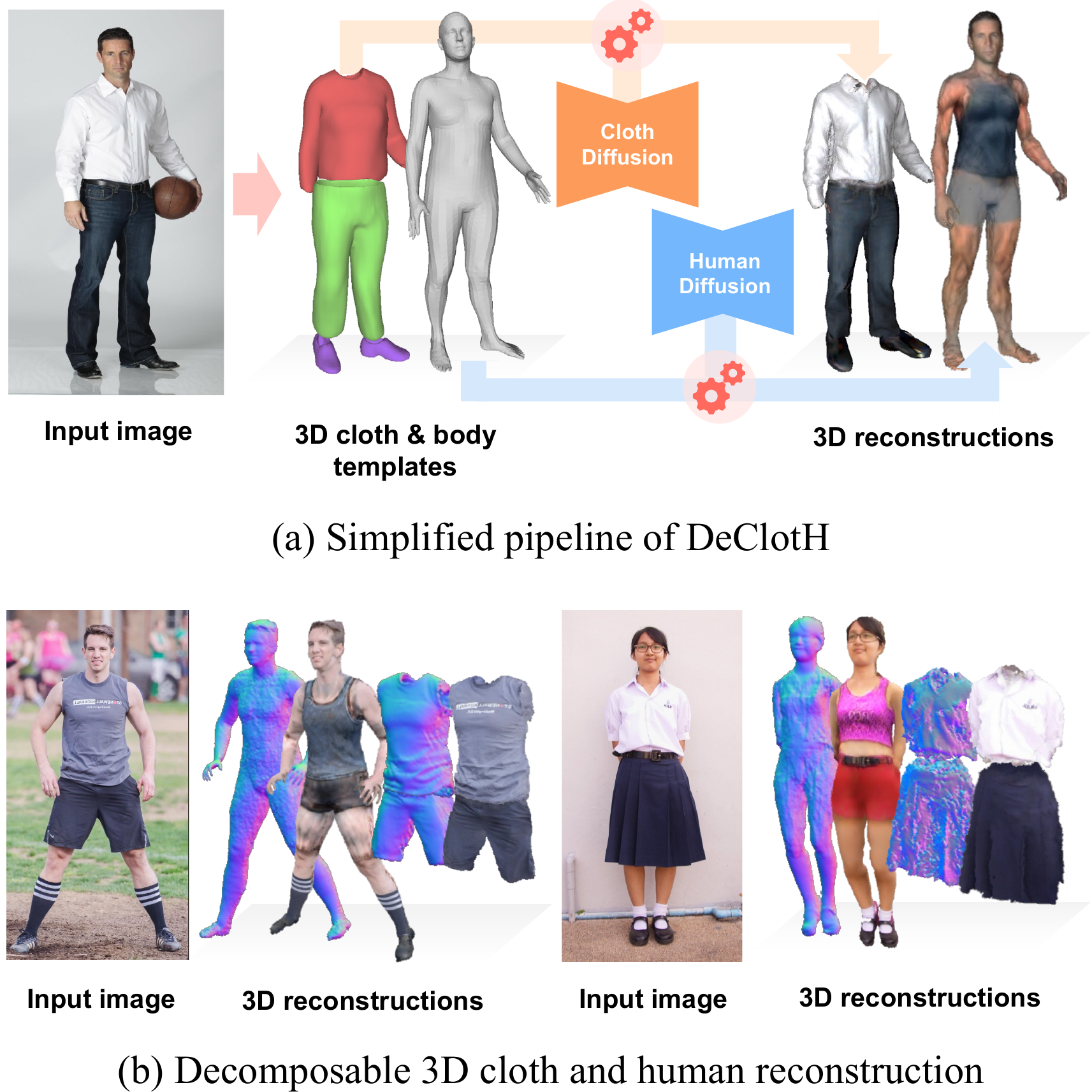}
  \vspace{-1.3em}
  \caption{\textbf{Overview of \ourmethod.}
  Given a single image, our framework reconstructs 3D cloth and human body based on the 3D cloth and body templates.
  }
  \vspace{-0.2em}
  \label{fig:teaser}
\end{figure}

Reconstructing 3D cloth and human body from a single image is an essential task for various applications including virtual try-on and AR/VR.
In recent years, numerous 3D clothed human reconstruction methods~\cite{albahar2023single,zhang2024humanref,ho2024sith,huang2024tech} have emerged with the advent of diffusion models~\cite{ho2020denoising}.
Although these methods achieve impressive reconstruction quality, they are inherently designed not to decompose the 3D cloth and human body, limiting their downstream applications.
In this regard, we tackle the more challenging task of separately reconstructing the 3D cloth and human body directly from a single image, as illustrated in \cref{fig:teaser}.

Despite the potential applications of decomposable 3D cloth and human body reconstruction, it has not been extensively explored.
One major problem is severe occlusion between the cloth and human body, with cloth covering substantial portions of the human body surface.
Such an occlusion makes it difficult to infer the overall geometry and texture of the invisible parts between 3D cloth and human body.
Additionally, image evidence (\textit{e.g.}, cloth silhouette) of the input image is often imperfect due to occlusion, leading to reconstruction that can overfit to the imperfect evidence.
For these reasons, the reconstruction of decomposable 3D cloth and human body is considerably more challenging than the previous tasks, which do not account for the occlusion between cloth and human body.

\begin{figure}[t!]
  \centering
  \includegraphics[width=1.0\linewidth]{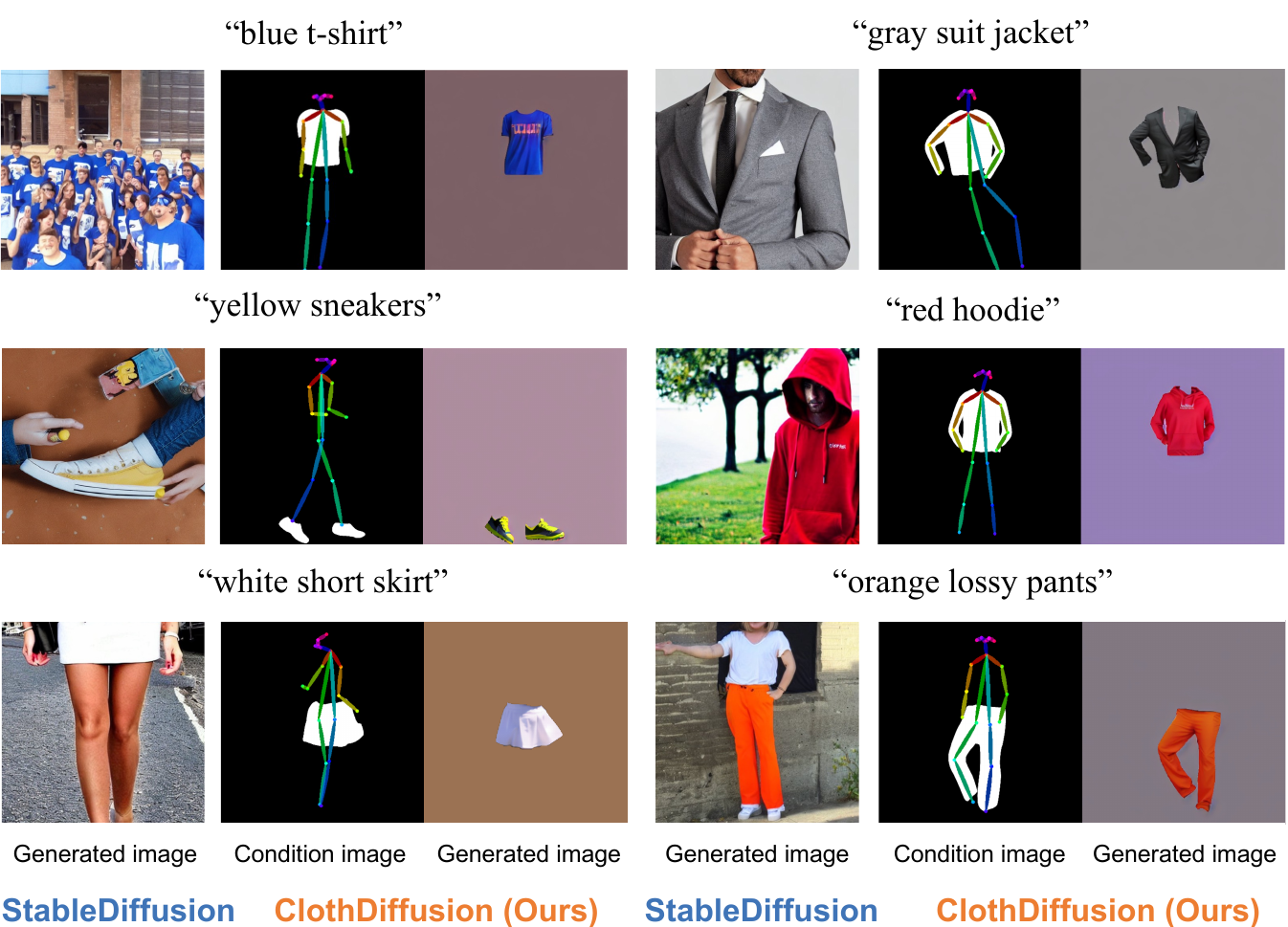}
  \vspace*{-1.7em}
  \caption{\textbf{Comparison between an existing diffusion model and ClothDiffusion.}
  Unlike the representative diffusion model, StableDiffusion~\cite{rombach2022high}, our ClothDiffusion generates cloth-specific images and can be controlled by cloth and human body templates.
  }
  \vspace*{-0.2em}
  \label{fig:intro_clothdiffusion}
\end{figure}

Recently, score distillation sampling (SDS)~\cite{poole2022dreamfusion} loss function has gained popularity in the 3D clothed human reconstruction literature~\cite{albahar2023single,zhang2024humanref,ho2024sith,huang2024tech} for inferring the geometry and texture of the occluded human parts.
The SDS loss enhances reconstruction quality by leveraging the image prior knowledge of a pre-trained text-to-image diffusion model.
Although employing the diffusion model is also promising for reconstructing 3D cloth, we observe that naively using such a strategy can provide incorrect guidance about cloth appearance.
As shown in \cref{fig:intro_clothdiffusion}, a representative diffusion model, StableDiffusion~\cite{rombach2022high}, typically generates images that focus on both cloth and human body, rather than the cloth itself.
Since its image prior knowledge contains a mixture of both cloth and human body, the diffusion model is unsuitable for reconstructing 3D cloth from the human body separately.
Furthermore, the generated images exhibit excessive diversity in scale, position, and cloth deformation, making it difficult to guide reconstruction of the actual cloth regions.

To address the above challenges, we present \textbf{\ourmethod} (\textbf{De}composable 3D \textbf{Clot}h and \textbf{H}uman body reconstruction), a template-based optimization framework designed for reconstructing 3D cloth and the human body from a single image.
This framework utilizes 3D template models of cloth and human body as strong geometric priors for reconstruction to mitigate erroneous results caused by occlusion.
The 3D template models represent the typical shapes of real-world clothes and human bodies by parameterizing them into a low-dimensional latent space.
For example, SMPLicit~\cite{corona2021smplicit} parameterizes 3D clothes with cloth style and looseness, and SMPL~\cite{loper2015smpl} parameterizes 3D human bodies with human pose and shape.
Based on these template models, we design a template regularization loss function, which constrains the reconstructed 3D cloth and human body to be close to their 3D template models.
Such constraints reduce heavy reliance on the imperfect image evidence caused by occlusions in the input image, leading to more plausible shapes for 3D cloth and human body.
Consequently, leveraging 3D template models of cloth and the human body, our framework produces robust reconstructions under severe occlusion between the cloth and human body.

Additionally, we devise a new diffusion model, ClothDiffusion, which overcomes the drawbacks of the existing diffusion model (\textit{i.e.}, StableDiffusion~\cite{rombach2022high}) for 3D cloth reconstruction.
Unlike StableDiffusion, which generates mixed content of cloth and human body, ClothDiffusion is specifically trained to generate only cloth images, as shown in \cref{fig:intro_clothdiffusion}.
This attribute of ClothDiffusion is highly beneficial for reconstructing 3D cloth, by providing prior image knowledge specialized for cloth geometry and texture.
Additionally, ClothDiffusion can be controlled by incorporating 3D template models as regional information for guidance.
From the 3D template models, we extract a cloth silhouette and a human skeleton and forward them to ClothDiffusion. 
By applying this regional information, ClothDiffusion can provide appropriate guidance that aligns with the actual cloth shape and human pose of the input image.
Thus, utilizing ClothDiffusion leads to the delicate geometry and texture of 3D cloth along with reconstructing the 3D human body.

Our extensive experiments demonstrate that \ourmethod~produces significantly more accurate reconstruction results than baseline methods, for both 3D cloth and human body. 
Our contributions can be summarized as follows.
\begin{itemize}
\item We present \ourmethod, which reconstructs a decomposable 3D cloth and human body from a single image, allowing various applications.
\item To address occlusion in reconstruction, we propose using 3D template models of the cloth and human body as beneficial constraints during reconstruction.
\item To improve the reconstruction of 3D cloth, we introduce a cloth diffusion model that provides contextual information specialized in cloth geometry and texture.
\end{itemize}

\section{Related works}
\label{sec:related_works}
\noindent\textbf{3D clothed human reconstruction.}
Most of the pioneering works~\cite{saito2019pifu,saito2020pifuhd,zheng2021pamir,xiu2022icon,xiu2023econ,han2023high,alldieck2022photorealistic,zheng2019deephuman,huang2020arch,he2021arch++,zhang2024global,pan2024humansplat,hu2023sherf,alldieck2022phorhum,corona2023structured,liao2023high,wang2023complete,pan2025humansplat} of 3D clothed human reconstruction train their networks based on 3D scan data, such as RenderPeople~\cite{renderpeople} and THuman2.0~\cite{tao2021function4d} datasets.
PIFuHD~\cite{saito2020pifuhd} introduced a coarse-to-fine framework to learn the high-resolution geometry of 3D clothed humans.
PHORHUM~\cite{alldieck2022photorealistic} photo-realistically reconstructs the 3D clothed humans while inferring shading.
ECON~\cite{xiu2023econ} proposed a method that combines human normal maps with a 3D parametric human body for fine geometric details. 
Recently, several works~\cite{albahar2023single,ho2024sith,huang2024tech,kolotouros2024avatarpopup} have leveraged a pre-trained text-to-image diffusion model~\cite{rombach2022high} to reconstruct the geometry and texture of 3D clothed humans, utilizing the strong prior knowledge of the diffusion model.
HumanSGD~\cite{albahar2023single} proposed a human mesh inpainting method with a shape-guided diffusion model.
SiTH~\cite{ho2024sith} presented a two-stage pipeline that predicts a back-view image and reconstructs a 3D clothed human based on the front- and back-view images.
TeCH~\cite{huang2024tech} utilized a Visual Question Answering (VQA) module to obtain descriptive text prompts from the image as input to the diffusion model.
The existing methods reconstruct the 3D clothed human as one unified mesh, which cannot be decomposed into the 3D cloth and human body components.
On the other hand, our \ourmethod~enables the separation of 3D cloth and human body from the reconstructions, allowing for a wide range of applications.

\noindent\textbf{3D template models.}
3D template models of cloth and human body are essential for recent 3D clothed human reconstruction~\cite{jiang2020bcnet,corona2021smplicit,zhu2022registering,moon20223d,de2023drapenet,bhatnagar2019multi} and 3D human body reconstruction~\cite{kanazawa2018end,kolotouros2019learning,kocabas2021pare,moon2022accurate,li2022cliff,kocabas2020vibe,choi2021beyond,choi2022learning,pymaf2021,nam2023cyclic,choi2023rethinking}.
These reconstruction methods predict the parameters of their respective 3D template models to reconstruct 3D clothes or human bodies.
BCNet~\cite{jiang2020bcnet} presented a 3D clothed human reconstruction system that predicts the PCA coefficients of 3D cloth template models.
ClothWild~\cite{moon20223d} leveraged a weakly supervised learning strategy for 3D clothed humans by using a 3D cloth template model, SMPLicit~\cite{corona2021smplicit}.
HMR~\cite{kanazawa2018end} proposed a 3D human body reconstruction framework with adversarial loss to learn plausible 3D pose and shape of the 3D human body template model, SMPL~\cite{loper2015smpl}.
PIXIE~\cite{feng2021collaborative} proposed a 3D human reconstruction method that estimates 3D hand pose and facial expression using SMPL-X~\cite{pavlakos2019expressive}.
These template-based reconstruction methods have a critical drawback because they are constrained by pre-defined template topology.
Due to this limitation, they generally reconstruct an overly smoothed mesh that does not capture the actual wrinkles of the clothes.
On the other hand, our framework covers fine details through the template-based optimization, integrating image evidence (\textit{e.g.}, cloth silhouettes) with 3D template models in the reconstruction.

\noindent\textbf{3D cloth decomposition.}
Encouraged by recent attention to 3D virtual try-on, several works~\cite{zielonka2023drivable,xiu2024puzzleavatar,kim2024gala,tiwari2020sizer,wang20244d} have proposed methods to decompose 3D clothes from a 3D human scan, multi-view images, or a video.
SIZER~\cite{tiwari2020sizer} introduced a method to segment 3D cloth meshes from a 3D human scan by voting on mesh vertices corresponding to cloth labels.
4D-DRESS~\cite{wang20244d} leveraged graph cut optimization to decompose 3D clothes from a 3D human scan.
GALA~\cite{kim2024gala} proposed a method to utilize a text-to-image diffusion model as a valuable prior to decomposition.
Compared to these methods, our framework tackles a much more challenging task, reconstructing decomposed 3D cloth and human body from a single image.

\section{\ourmethod}
\label{sec:proposed_method}
\begin{figure*}[t!]
  \centering
  \includegraphics[width=1.0\linewidth]{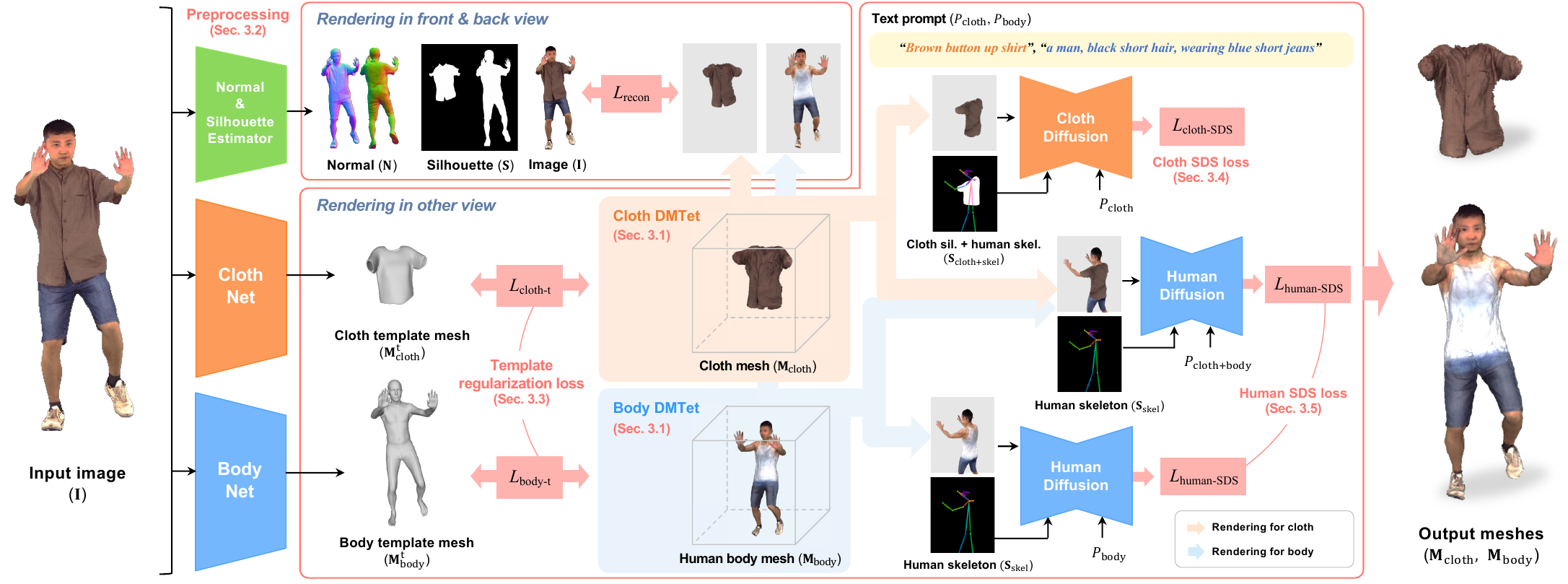}
  \vspace*{-1.4em}
  \caption{\textbf{Overall pipeline of \ourmethod.}
  Given an input image $\textbf{I}$, \ourmethod~optimizes 3D cloth and human body, represented by DMTets (\cref{sec:dmtet}).
  For the optimization, we extract normal map $\textbf{N}$, silhouette $\textbf{S}$, and 3D template meshes ($\textbf{M}^{\text{t}}_{\text{cloth}}$ and $\textbf{M}^{\text{t}}_{\text{body}}$) (\cref{sec:preprocessing}).
  Subsequently, the 3D cloth and human body are optimized by three core loss functions: template regularization loss (\cref{sec:regloss}), cloth SDS loss (\cref{sec:cloth_sds}), and human SDS loss (\cref{sec:human_sds}).
  }
   \vspace*{-0.2em}
  \label{fig:overall_pipeline}
\end{figure*}
Fig.~\ref{fig:overall_pipeline} illustrates the overall pipeline of our \ourmethod.
Given an input image $\mathbf{I}$ and a target cloth type $C$, our framework optimizes both the target 3D cloth mesh $\mathbf{M}_{\text{cloth}}$ and the 3D human mesh $\mathbf{M}_{\text{body}}$ not wearing the 3D cloth.
To simplify the description, we refer to the 3D human without the target cloth as the ``human body''.
In the following sections, we first describe the 3D geometry representation (\cref{sec:dmtet}) and image preprocessing (\cref{sec:preprocessing}) for the optimization of 3D cloth and human body.
Subsequently, we provide detailed descriptions of three core loss functions in our framework: template regularization loss (\cref{sec:regloss}), cloth SDS loss (\cref{sec:cloth_sds}), and human SDS loss (\cref{sec:human_sds}).
Finally, we explain the overall optimization process (\cref{sec:optimization}).

\subsection{3D geometric representation (DMTet)}
\label{sec:dmtet}
In our framework, Deep Marching Tetrahedra~\cite{shen2021deep} (DMTet) is utilized as the 3D geometric representation of 3D cloth and human body.
DMTet represents 3D geometry with a deformable tetrahedral grid ($\mathbf{X}_T$, $T$), where $\mathbf{X}_T$ denotes 3D vertices of the tetrahedral grid and $T$ defines the tetrahedral structure.
Specifically, the MLP network of DMTet predicts the signed distance value from the 3D geometry surface for each vertex of the grid.
We adopt two DMTets to represent 3D cloth and human body in a canonical pose (A-pose), respectively.
To obtain 3D cloth mesh $\mathbf{M}_{\text{cloth}}$ and 3D human body mesh $\mathbf{M}_{\text{body}}$ from the DMTets, we extract meshes via the Marching Tetrahedra~\cite{doi1991efficient} (MT) algorithm and transform the meshes via a linear blend skinning (LBS), which is pre-defined in the SMPL+H~\cite{loper2015smpl} human model.

\subsection{Image preprocessing}
\label{sec:preprocessing}
To optimize DMTets of 3D cloth and human body, we gather multiple optimization targets: normal map $\mathbf{N}$, silhouette $\mathbf{S}$, and 3D template meshes ($\mathbf{M}^{\text{t}}_{\text{cloth}}$ and $\mathbf{M}^{\text{t}}_{\text{body}}$).

\noindent\textbf{Normal \& silhouette estimator.}
The normal maps $\mathbf{N}$ of the front and back views are obtained by the normal estimator of ECON~\cite{xiu2023econ} from the input image.
The silhouettes $\mathbf{S}$ of cloth and human are acquired by running the off-the-shelf segmentation method, SAM~\cite{kirillov2023segment}, given input image $\mathbf{I}$ and text prompt ($\mathbf{t}_{\text{cloth}}$, $\mathbf{t}_{\text{human}}$).

\noindent\textbf{ClothNet.}
ClothNet predicts a 3D cloth template mesh $\mathbf{M}^{\text{t}}_{\text{cloth}}$ of the target cloth from the input image.
We use ClothWild~\cite{moon20223d} as ClothNet, which achieves state-of-the-art performance on in-the-wild images.

\noindent\textbf{BodyNet.}
BodyNet predicts the 3D body template mesh $\mathbf{M}^{\text{t}}_{\text{body}}$, by estimating the 3D human pose and shape of the SMPL + H model~\cite{loper2015smpl} from the input image.
We modify PIXIE~\cite{feng2021collaborative} to infer SMPL+H and use it as BodyNet.

\subsection{Template regularization loss}
\label{sec:regloss}
The template regularization losses, $L_{\text{cloth-t}}$ and $L_{\text{body-t}}$, enforce optimized 3D cloth and human body meshes to be close to their 3D template meshes ($\mathbf{M}^{\text{t}}_{\text{cloth}}$ and $\mathbf{M}^{\text{t}}_{\text{body}}$).
The loss functions are defined as 
\begin{align}
L_{\text{cloth-t}} = \lVert \mathcal{R}_{\text{sil}}(\mathbf{M_{\text{cloth}}}, \mathbf{k}) - \mathcal{R}_{\text{sil}}(\mathbf{M^{\text{t}}_{\text{cloth}}}, \mathbf{k}) \rVert_2, \\
L_{\text{body-t}} = \lVert \mathcal{R}_{\text{sil}}(\mathbf{M_{\text{body}}}, \mathbf{k}) - \mathcal{R}_{\text{sil}}(\mathbf{M^{\text{t}}_{\text{body}}}, \mathbf{k}) \rVert_2,
\end{align}
where $\mathcal{R}_{\text{sil}}$ is a silhouette renderer, and $\mathbf{k}$ is a camera parameter for the rendering.
These losses constrain the projected 3D meshes to be close to the projection of their 3D template meshes.
ClothNet and BodyNet, which produce the 3D template meshes, are trained on in-the-wild datasets containing various images with occlusions. 
The 3D template meshes exhibit high robustness against occlusions, leveraging learned data-driven knowledge from the in-the-wild datasets.
Consequently, the template regularization loss prevents the erroneous reconstruction of 3D cloth and human body from occlusion.

\subsection{Cloth SDS loss}
\label{sec:cloth_sds}
Cloth SDS loss $L_{\text{cloth-SDS}}$ supervises the geometry and texture of 3D cloth, particularly for regions that are not visible in the front view.
Our cloth SDS loss follows the basic formulation of SDS loss proposed by DreamFusion~\cite{poole2022dreamfusion}.
Given a 3D mesh parameterized with $\phi$, the SDS loss updates $\phi$ based on the rendered image $\mathbf{x}$ of the 3D mesh, image condition $\mathbf{c}$, and text prompt $P$ for the diffusion model.
The gradient of the SDS loss is calculated as
\begin{equation}
\begin{split}
\nabla_{\phi} L_{\text{SDS}}(\mathbf{x}, \mathbf{c}, P) = \mathbb{E}[w_t (\hat{\epsilon}(\mathbf{x}_{t}; \mathbf{c}, P, t) - \epsilon) \frac{\partial \mathbf{x}_{t}}{\partial \phi}],
\end{split}
\end{equation}
where $t$ denotes a noise level, $\mathbf{x}_{t}$ is a rendered image with noise, and $w_{t}$ is a weighting variable dependent on the noise level $t$.
This loss function computes the distance between the predicted noise $\hat{\epsilon}(\cdot)$ and the sampled noise $\epsilon$ from the diffusion model.
Accordingly, the SDS loss guides the 3D mesh rendering to follow a visually coherent appearance that aligns with the image condition and text prompt.

Unlike the basic SDS loss, the cloth SDS loss function exploits a new diffusion model, ClothDiffusion, instead of StableDiffusion~\cite{rombach2022high}.
ClothDiffusion is trained to generate cloth-specific images using cloth silhouettes and human skeletons as conditioning inputs.
Cloth SDS loss supervises the rendered normal maps and RGB images as follows:
\begin{align}
L^{\text{norm}}_{\text{cloth-SDS}} = L_{\text{SDS}}(\mathbf{N}^{\mathbf{k}}_{\text{cloth}}, \mathbf{S}^{\mathbf{k}}_{\text{cloth+skel}}, P_{\text{cloth}}), \\
L^{\text{rgb}}_{\text{cloth-SDS}} = L_{\text{SDS}}(\mathbf{I}^{\mathbf{k}}_{\text{cloth}}, \mathbf{S}^{\mathbf{k}}_{\text{cloth+skel}}, P_{\text{cloth}}), 
\end{align}
where $\mathbf{N}^{\mathbf{k}}_{\text{cloth}}$ and $\mathbf{I}^{\mathbf{k}}_{\text{cloth}}$ are the normal map and the RGB image rendered from the 3D cloth mesh $\mathbf{M}_{\text{cloth}}$ with a camera parameter $\mathbf{k}$.
The $\mathbf{S}^{\mathbf{k}}_{\text{cloth+body}}$ is a combination of two silhouettes of 3D cloth template mesh $\mathbf{M}^{\text{t}}_{\text{cloth}}$ and 3D human skeleton extracted from the 3D body template mesh $\mathbf{M}^{\text{t}}_{\text{body}}$.
The cloth SDS loss provides rich contextual information on cloth appearances using cloth-specific prior knowledge of ClothDiffusion.
Additionally, cloth SDS loss can accurately guide the reconstruction of cloth regions by utilizing cloth silhouettes and human skeletons as regional information.

\subsection{Human SDS loss}
\label{sec:human_sds}
Human loss of SDS $L_{\text{human-SDS}}$ supervises the geometry and texture of the occluded parts of both the 3D cloth and the human body.
This loss is based on HumanDiffusion, another diffusion model that generates human images conditioned on human skeletons.
We adopt pre-trained weights of ControlNet~\cite{zhang2023adding} for the HumanDiffusion.
Human SDS loss function supervises the rendered normal map and RGB image as follows:
\begin{equation}
\begin{split}
L^{\text{norm}}_{\text{human-SDS}} = L_{\text{SDS}}(\mathbf{N}^{\mathbf{k}}_{\text{body}}, \mathbf{S}^{\mathbf{k}}_{\text{skel}}, P_{\text{body}}) \qquad\qquad \\ 
\qquad\qquad+ L_{\text{SDS}}(\mathbf{N}^{\mathbf{k}}_{\text{cloth+body}}, \mathbf{S}^{\mathbf{k}}_{\text{skel}}, P_{\text{cloth+body}}),
\end{split}
\end{equation}
\vspace*{-0.2em}
\begin{equation}
\begin{split}
L^{\text{rgb}}_{\text{human-SDS}} = L_{\text{SDS}}(\mathbf{I}^{\mathbf{k}}_{\text{body}}, \mathbf{S}^{\mathbf{k}}_{\text{skel}}, P_{\text{body}}) \qquad\qquad \\ 
\qquad\qquad+ L_{\text{SDS}}(\mathbf{N}^{\mathbf{k}}_{\text{cloth+body}}, \mathbf{S}^{\mathbf{k}}_{\text{skel}}, P_{\text{cloth+body}}),
\end{split}
\end{equation}
where $\mathbf{N}^{\mathbf{k}}_{\text{body}}$ and $\mathbf{I}^{\mathbf{k}}_{\text{body}}$ are the normal map and the RGB image rendered from the 3D human body mesh $\mathbf{M}_{\text{body}}$.
Further, $\mathbf{N}^{\mathbf{k}}_{\text{cloth+body}}$ and $\mathbf{I}^{\mathbf{k}}_{\text{cloth+body}}$ are rendered from composition of the 3D cloth and human body meshes ($\mathbf{M}_{\text{cloth}}$ and $\mathbf{M}_{\text{body}}$).
$\mathbf{S}^{\mathbf{k}}_{\text{skel}}$ is the rendered image from 3D human skeleton extracted from the 3D body template mesh $\mathbf{M}^{\text{t}}_{\text{body}}$.
By incorporating human skeletons, the human SDS loss provides accurate guidance for human regions.

\subsection{Optimization procedure}
\label{sec:optimization}
Based on the above loss functions, we optimize the 3D cloth mesh $\mathbf{M}_{\text{cloth}}$ and the 3D human body mesh $\mathbf{M}_{\text{body}}$ through two stages: geometry stage and texture stage.

\noindent\textbf{Geometry stage.}
In the geometry stage, we optimize the DMTets of 3D cloth and human body by minimizing loss function $L_{\text{geo}}$, defined as follows:
\begin{equation}
\begin{split}
L_{\text{geo}} = L_{\text{cloth-t}} + L_{\text{body-t}} + L^{\text{norm}}_{\text{cloth-SDS}} + L^{\text{norm}}_{\text{human-SDS}}  + L^{\text{geo}}_{\text{recon}}.
\end{split}
\end{equation}
$L^{\text{geo}}_{\text{recon}}$ is defined as
\begin{equation}
\begin{split}
L^{\text{geo}}_{\text{recon}} &= L_{\text{normal}} + L_{\text{sil}} + L_{\text{pen}}.
\end{split}
\end{equation}
$L_{\text{normal}}$ is the L2 distance between rendered normal maps and their optimization targets $\mathbf{N}$ in the front and back view.
$L_{\text{sil}}$ is the L2 distance between the rendered silhouettes and their optimization targets $\mathbf{S}$ in the front and back view.
$L_{\text{normal}}$ and $L_{\text{sil}}$ are calculated for both the cloth and the human body.
$L_{\text{pen}}$ penalizes intersection between the 3D cloth and human body meshes.

\noindent\textbf{Texture stage.}
In the texture stage, we optimize the texture of the meshes ($\mathbf{M}_{\text{cloth}}$ and $\mathbf{M}_{\text{body}}$) obtained from the geometry stage.
To this end, we construct MLP networks ($\Psi_{\text{cloth}}$ and $\Psi_{\text{human}}$) that predict RGB color given the vertex coordinate of the mesh.
These MLP networks are trained by minimizing the loss function $L_{\text{tex}}$, defined as:
\begin{equation}
\begin{split}
L_{\text{tex}} &= L^{\text{rgb}}_{\text{cloth-SDS}} + L^{\text{rgb}}_{\text{human-SDS}} + L^{\text{tex}}_{\text{recon}}.
\end{split}
\end{equation}
$L^{\text{tex}}_{\text{recon}}$ is a combination of L2 and LPIPS~\cite{zhang2018unreasonable} distances between rendered RGB images and their optimization target (\textit{i.e.}, $\mathbf{I}$) in the front and back view.
$L^{\text{tex}}_{\text{recon}}$ is calculated for both cloth and human body.

\begin{figure}[t!]
  \centering
  \includegraphics[width=1.0\linewidth]{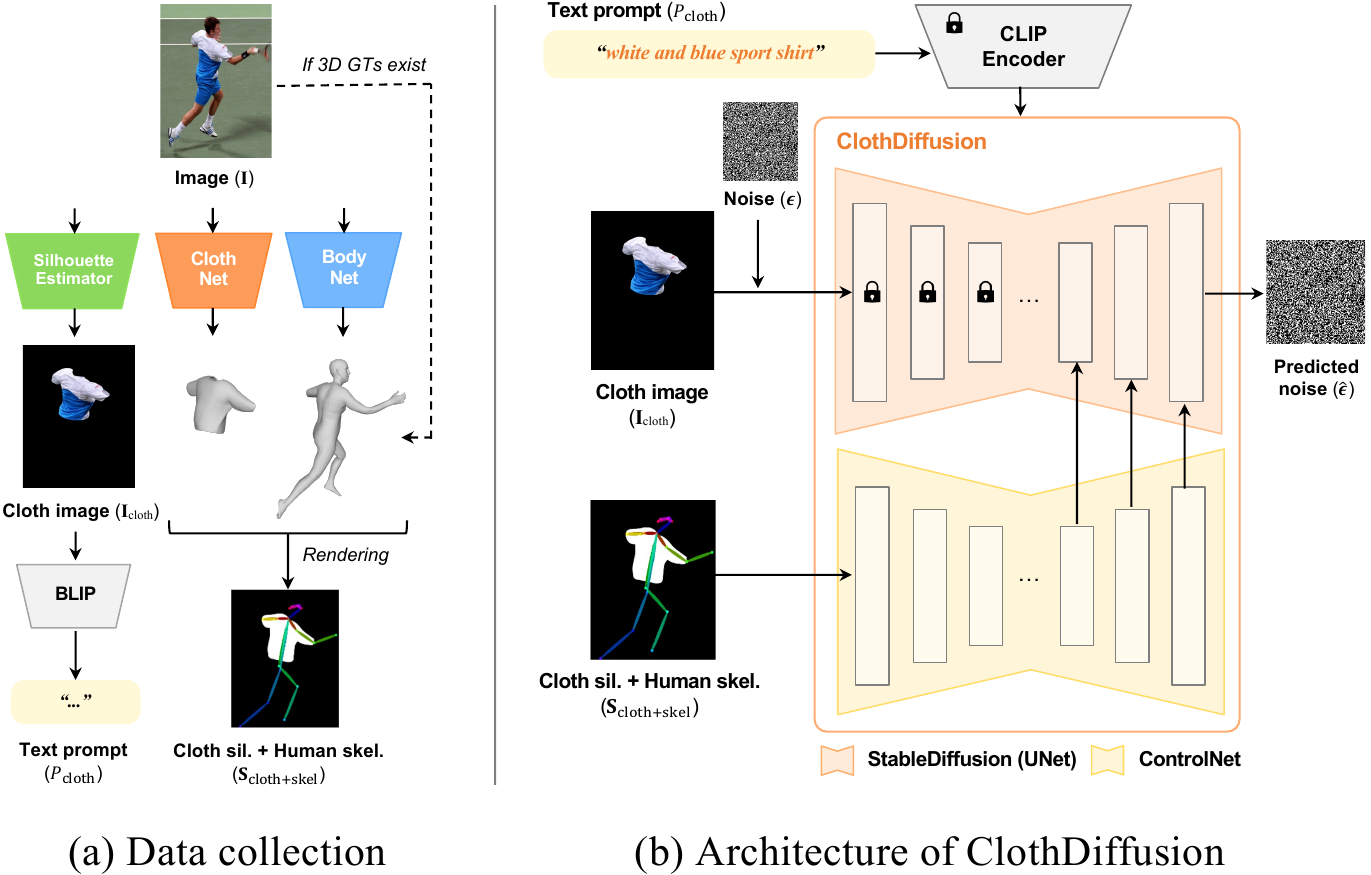}
  \vspace*{-1.5em}
  \caption{\textbf{Training process of ClothDiffusion.}
  We train the ClothDiffusion based on our collected cloth-specific training data.
  The ClothDiffusion follows ControlNet architecture with the pre-trained StableDiffusion.
  }
  \vspace*{-0.2em}
  \label{fig:train_clothdiffusion}
\end{figure}
\section{ClothDiffusion}
\label{sec:clothdiffusion}
\cref{fig:train_clothdiffusion} shows the training process of ClothDiffusion, which is used in the cloth SDS loss.

\subsection{Training data collection}
To train ClothDiffusion, we collect three data types: 1) cloth images, 2) conditional images for generation, and 3) text prompts.
The cloth images are obtained by running the silhouette estimator~\cite{kirillov2023segment} from images of the training datasets.
The conditional images are acquired by projecting a 3D cloth template mesh and a human skeleton extracted from a 3D body mesh.
The 3D cloth and human templates are derived from an image using ClothNet and BodyNet unless 3D ground-truths (GTs) are available in the training datasets.
The text prompts are acquired from BLIP~\cite{li2022blip}, a state-of-the-art image captioning method.
The acquired text prompt provides detailed descriptions of cloth color, shape, and style in the cloth image.

\subsection{Learning cloth generation}
ClothDiffusion consists of two networks: StableDiffusion~\cite{rombach2022high} and ControlNet~\cite{zhang2023adding}.
StableDiffusion estimates noise, given a latent embedding of a cloth image with sampled noise.
ControlNet takes a conditional image (\textit{i.e.}, cloth silhouette and human skeleton) and gives additional information for StableDiffusion to generate a plausible image based on the conditional image.
Using the pre-trained weights of StableDiffusion, we train ControlNet and partial layers of StableDiffusion while keeping its other layers frozen, following the fine-tuning strategy of ControlNet.
This fine-tuning strategy allows ClothDiffusion to generate realistic cloth images by leveraging the strong prior knowledge embedded in the pre-trained StableDiffusion.
\section{Experiments}
\label{sec:experiments}
\subsection{Datasets}
\noindent\textbf{4D-DRESS.}
4D-DRESS~\cite{wang20244d} contains high-quality 3D scans of 64 clothing sequences, with diverse human poses.
This dataset is used solely for evaluation purposes.
For each clothing sequence in 4D-DRESS, we randomly sample one human pose and render the corresponding 3D scan from pre-defined camera viewpoints to obtain the test images.
The evaluation set includes 64 test images, and 124 clothes appear in the test images.

\noindent\textbf{THuman2.0.}
THuman2.0~\cite{tao2021function4d} contains 3D scans of 525 clothed humans.
From all scans, we uniformly sample 52 scans for evaluation. 
As THuman2.0 does not contain GTs of 3D cloth, we generate 3D cloth pseudo-GTs by running GALA~\cite{kim2024gala} on the 3D human scans.
We obtain test images by rendering the 3D scans with pre-defined camera viewpoints.
The evaluation set includes 52 test images, and 104 clothes appear in the test images.

\noindent\textbf{Training datasets of ClothDiffusion.}
DeepFashion~\cite{liu2016deepfashion}, SHHQ~\cite{fu2022stylegan}, MSCOCO~\cite{lin2014microsoft}, and THuman2.0~\cite{tao2021function4d} are used to train ClothDiffusion.
DeepFashion~\cite{liu2016deepfashion} and SHHQ~\cite{fu2022stylegan} are large-scale datasets that contain diverse images of clothed humans.
We use the official training set from DeepFashion and uniformly sample 10,000 images from SHHQ.
MSCOCO~\cite{lin2014microsoft} contains a wide variety of human poses, and we utilize its official training set.
THuman2.0~\cite{tao2021function4d} provides multi-view images of clothed people, excluding duplicates from the evaluation set.

\subsection{Evaluation metrics}
\noindent\textbf{3D geometry reconstruction.}
We evaluate the geometry of 3D reconstructions by measuring CD (chamfer distance) and NC (normal consistency), following Huang~\etal~\cite{huang2024tech}.
Specifically, we apply Procrustes alignment on the reconstructed 3D meshes based on the SMPL-X meshes of 3D GTs.
With the aligned 3D meshes, we measure CD and NC between reconstructed and GT meshes.
To measure NC, we calculate the average L2 distance between normal images of reconstructed and GT meshes, rendered at \{0$^{\circ}$, 90$^{\circ}$, 180$^{\circ}$, 270$^{\circ}$\} from fixed viewpoints.

\noindent\textbf{3D texture reconstruction.}
We evaluate the texture of 3D reconstructions, using PSNR (peak signal-to-noise ratio) and LPIPS~\cite{zhang2018unreasonable}. (learned perceptual image patch similarity).
Before evaluation, we align the reconstructions with 3D human GT poses to eliminate the influence of geometric errors.
Then, we calculate PSNR and LPIPS between RGB images of reconstructed and GT meshes rendered at \{0$^{\circ}$, 90$^{\circ}$, 180$^{\circ}$, 270$^{\circ}$\} as in the NC measurement.

\noindent\textbf{2D image generation of diffusion model.}
To verify the feasibility of our proposed ClothDiffusion, we evaluate the generated images from the diffusion model via CLIP-Score~\cite{hessel2021clipscore} and Cloth-IoU.
CLIP-Score~\cite{hessel2021clipscore} measures the correlation between cloth text prompts and generated images.
Cloth-IoU measures a proportion of the intersection between generated cloth images and GT counterparts.
Specifically, we extract cloth silhouettes from the generated images by running SAM~\cite{kirillov2023segment} with cloth text prompts.
Then, we calculate the IoU between the extracted silhouettes and GT counterparts.
The cloth text prompts for evaluation are obtained by running BLIP~\cite{li2022blip} from the test images.

\begin{figure}[t!]
  \centering
  \includegraphics[width=1.0\linewidth]{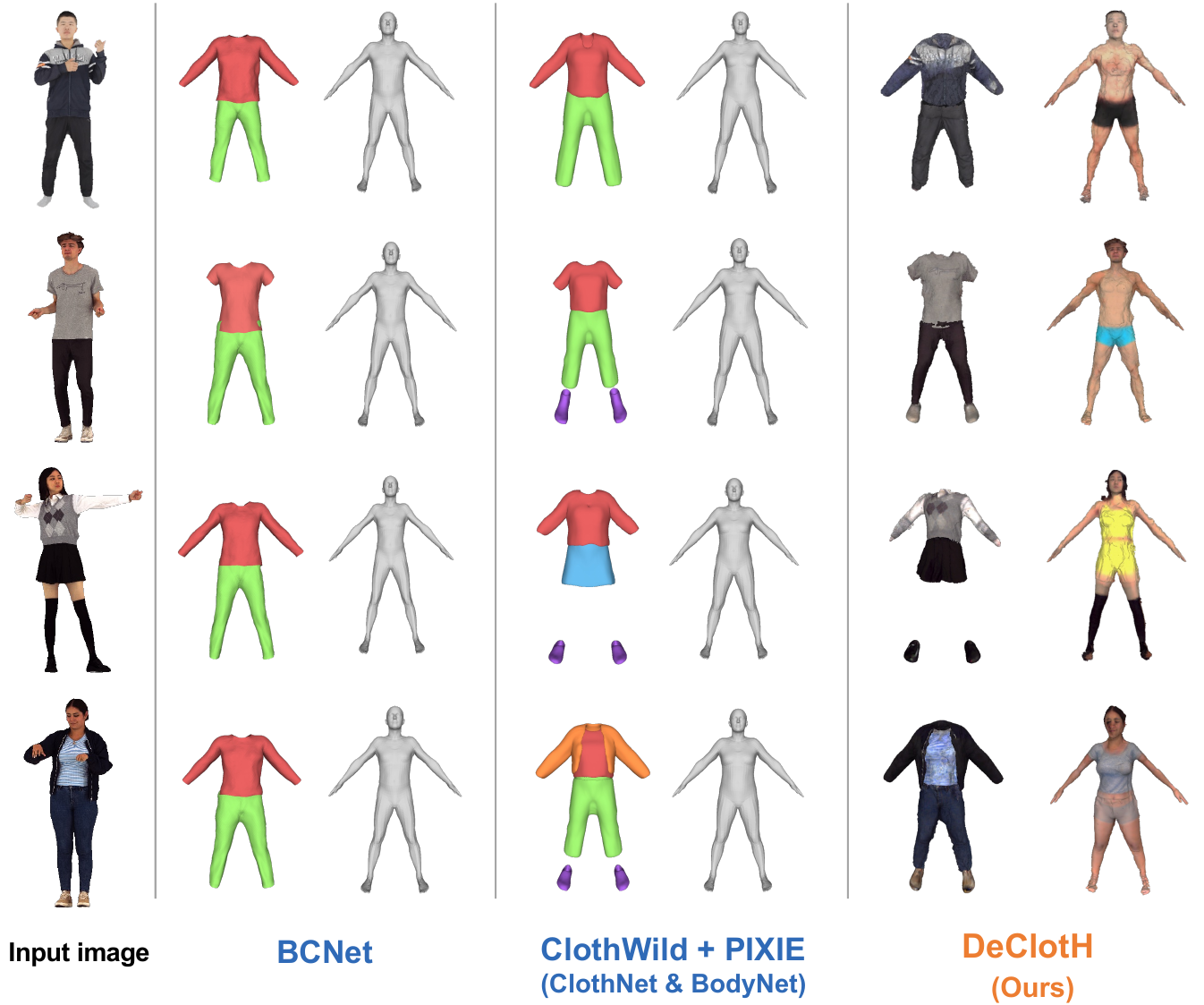}
  \vspace*{-1.6em}
  \caption{\textbf{Effects of the optimization process of~\ourmethod.} 
  }
   \vspace*{-0.4em}
  \label{fig:abla_optimization}
\end{figure}
\begin{table}[t]
\def\arraystretch{1.6}
\renewcommand{\tabcolsep}{0.8mm}
\footnotesize
\begin{center}
\scalebox{0.725}{
    \begin{tabular}{>{\raggedright\arraybackslash}m{3.5cm}|>{\centering\arraybackslash}m{1.75cm}>{\centering\arraybackslash}m{1.75cm}|>{\centering\arraybackslash}m{1.75cm}>{\centering\arraybackslash}m{1.75cm}}
    \specialrule{.1em}{.05em}{0.0em}
        &  \multicolumn{2}{c|}{4D-DRESS (cloth)} & \multicolumn{2}{c}{4D-DRESS (cloth+human)} \\
        Methods & CD$^{\downarrow}$ & NC$^{\downarrow}$ & CD$^{\downarrow}$ & NC$^{\downarrow}$ \\
        \hline
        BCNet~\cite{jiang2020bcnet} &4.387 & 0.046 & 3.925 & 0.090 \\
        SMPLicit~\cite{corona2021smplicit} &4.080 & 0.038 & 3.605 & 0.091 \\
        \vspace{1.5mm}\makecell[l]{ ClothWild~\cite{moon20223d} + PIXIE~\cite{feng2021collaborative} \\(ClothNet $\&$ BodyNet)}  & 4.100 & 0.038 & 3.526 & 0.087 \\
        \textbf{DeClotH (Ours)} & \textbf{3.902} & \textbf{0.037} & \textbf{3.292} & \textbf{0.079}  \\
        \specialrule{.1em}{-0.05em}{-0.05em}
    \end{tabular}   
}
\end{center}
    \vspace*{-1.3em}
    \caption{
        \textbf{Effectiveness of the \ourmethod's optimization process compared to 3D template meshes of ClothNet and BodyNet.
        }
    }
    \vspace*{+0.2em}
    \label{tab:abla_optimization}
\end{table}

\subsection{Ablation study}
\label{sec:ablation}
We carry out the ablation study on 4D-DRESS~\cite{wang20244d}.
As 4D-DRESS does not contain 3D human body scans excluding clothes, we compare methods in two tracks: evaluating 3D cloth (cloth) and evaluating the composition of 3D cloth and the human body (cloth $+$ human).

\noindent\textbf{Effectiveness of template-based optimization.}
\cref{fig:abla_optimization} and \cref{tab:abla_optimization} show that our optimization framework produces significantly more realistic 3D reconstruction results, compared to the 3D template meshes from ClothNet and BodyNet.
ClothNet and BodyNet are model-based methods that estimate the shape of pre-defined 3D template models of cloth and human body.
Thereby, their 3D template meshes cannot deviate from the topology of the template models.
Our framework, \ourmethod~builds upon these 3D template meshes by optimizing the 3D cloth and human body with the optimization targets (\textit{e.g.}, normal maps and silhouettes), which have rich geometric information in the input image.
Through the optimization process, our framework can reconstruct fine geometric details, such as cloth wrinkles, while keeping the overall shape of 3D templates.
Furthermore, our framework reconstructs the texture of cloth and human body along with the geometry, resulting in more lifelike reconstructions.

\begin{figure}[t]
  \centering
  \includegraphics[width=1.0\linewidth]{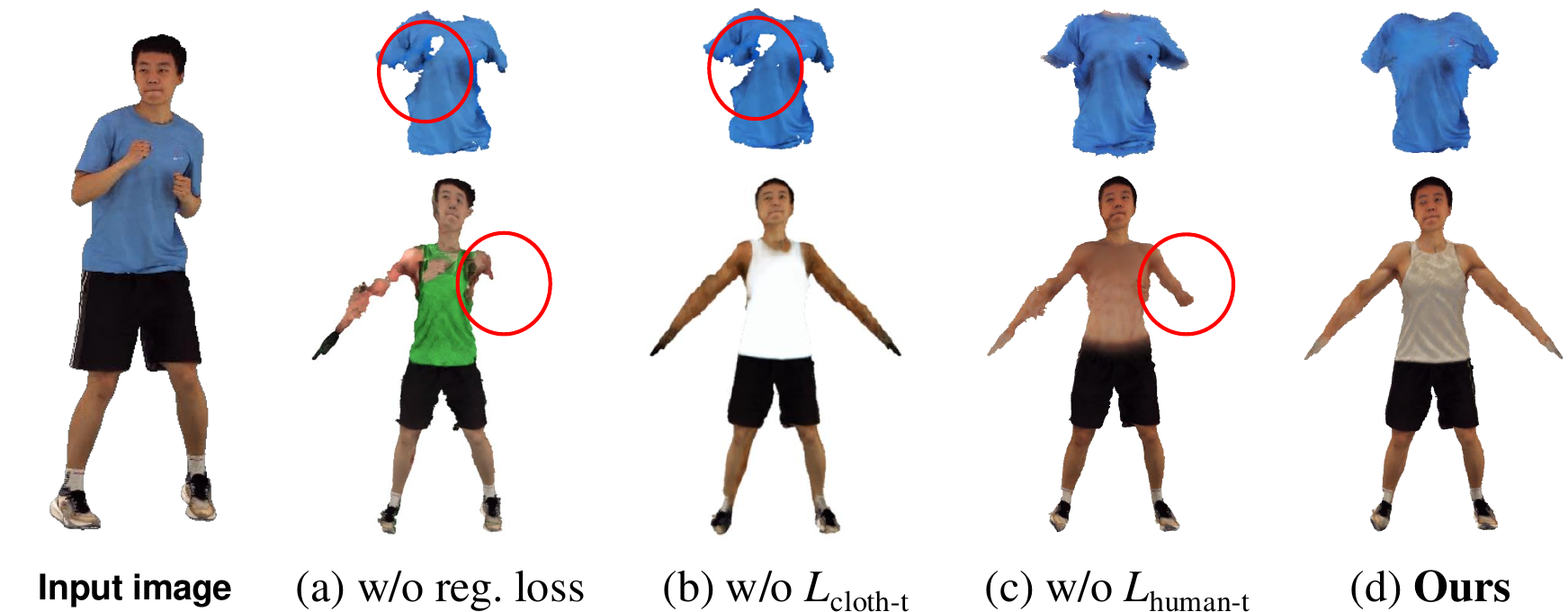}
  \vspace*{-1.4em}
  \caption{\textbf{Effects of template regularization loss.}
  }
  \vspace*{-0.0em}
  \label{fig:ablation_regloss}
\end{figure}
\begin{table}[t]
\def\arraystretch{1.48}
\renewcommand{\tabcolsep}{0.8mm}
\footnotesize
\begin{center}
\scalebox{0.744}{
    \begin{tabular}{>{\centering\arraybackslash}m{1.15cm}>{\centering\arraybackslash}m{1.15cm}|>{\centering\arraybackslash}m{0.8cm}>{\centering\arraybackslash}m{0.8cm}>{\centering\arraybackslash}m{0.94cm}>{\centering\arraybackslash}m{0.94cm}|>{\centering\arraybackslash}m{0.8cm}>{\centering\arraybackslash}m{0.8cm}>{\centering\arraybackslash}m{0.94cm}>{\centering\arraybackslash}m{0.94cm}}
    \specialrule{.1em}{.05em}{0.0em}
        & &  \multicolumn{4}{c|}{4D-DRESS (cloth)}  & \multicolumn{4}{c}{4D-DRESS (cloth + human)} \\
        $L_{\text{cloth-t}}$ & $L_{\text{body-t}}$ & CD$^{\downarrow}$ & NC$^{\downarrow}$ & PSNR$^{\uparrow}$ & LPIPS$^{\downarrow}$ & CD$^{\downarrow}$ & NC$^{\downarrow}$ & PSNR$^{\uparrow}$ & LPIPS$^{\downarrow}$ \\
        \hline
        \xmark & \xmark & 8.245 & 0.047 & 28.959 & 0.051 & 3.880 & 0.093 & 22.992 & 0.071 \\
        \xmark & \cmark & 8.386 & 0.047 & 29.028 & 0.055 & 3.567 & 0.084 & 23.149 & 0.070 \\
        \cmark & \xmark & 4.078 & 0.038 & 30.828 & 0.038 & 3.586 & 0.085 & 23.422 & 0.067 \\
        \cmark & \cmark & \textbf{3.902} & \textbf{0.037} & \textbf{31.582} & \textbf{0.033} & \textbf{3.292} & \textbf{0.079} & \textbf{23.921} & \textbf{0.065} \\
        \specialrule{.1em}{-0.05em}{-0.05em}
    \end{tabular}
}
\end{center}
    \vspace*{-1.5em}
    \caption{
    \textbf{Ablation studies for template regularization loss.}
    }
    \vspace*{-0.6em}
    \label{tab:abla_regloss}
\end{table}

\noindent\textbf{Effectiveness of template regularization loss.}
\cref{fig:ablation_regloss} and \cref{tab:abla_regloss} show that the template regularization loss significantly drops the 3D reconstruction error, especially when the cloth and human body are occluded.
Without the template regularization loss, the optimization of 3D cloth and human body highly relies on the optimization targets (\textit{e.g.}, cloth silhouette) of the front views.
Our proposed template regularization loss reduces the heavy reliance, by using the 3D template models as helpful supervision for the occluded parts.
Thus, the template regularization loss alleviates the occlusion issue and enhances the 3D reconstruction in both the cloth and the human body.

\begin{figure}[t!]
  \centering
  \includegraphics[width=1.0\linewidth]{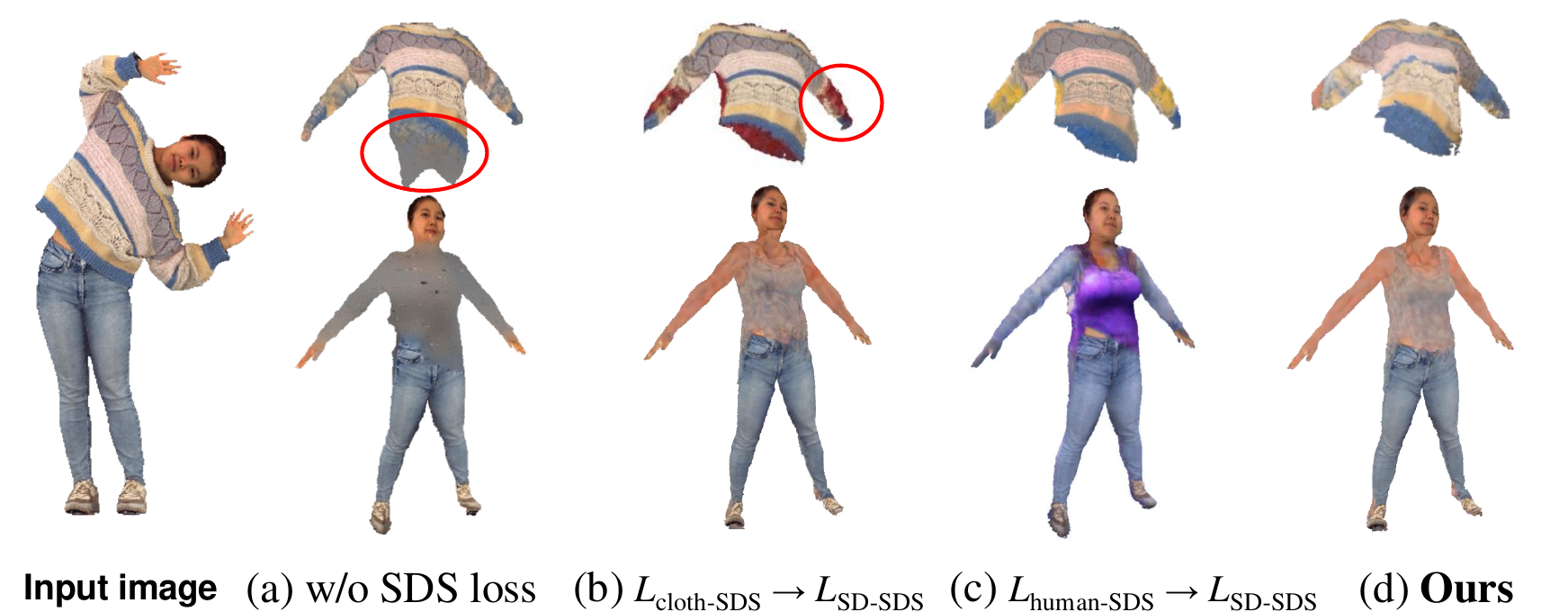}
  \vspace*{-1.3em}
  \caption{\textbf{Effects of cloth and human SDS loss.}
  }
  \vspace*{+0.1em}
  \label{fig:ablation_sdsloss}
\end{figure}

\noindent\textbf{Effectiveness of cloth \& human SDS loss.}
\cref{fig:ablation_sdsloss} and \cref{tab:abla_sdsloss} show that the cloth and human SDS losses are much more effective in reconstructing geometry and texture than vanilla SDS loss (\textit{i.e.}, $L_{\text{SD-SDS}}$) that uses StableDiffusion for reconstruction guidance.
As shown in \cref{fig:ablation_sdsloss} (c), the reconstructed 3D jacket has an artifact near the boundary between cloth and human body, when using the vanilla SDS loss.
This artifact indicates that StableDiffusion has inadequate prior knowledge to separately reconstruct 3D cloth from the human body.
On the other hand, the cloth SDS loss using ClothDiffusion effectively supervises the 3D geometry to follow the desired cloth shape.

\begin{table}[t]
\def\arraystretch{1.48}
\renewcommand{\tabcolsep}{0.8mm}
\footnotesize
\begin{center}
\scalebox{0.725}{
    \begin{tabular}{>{\centering\arraybackslash}m{1.38cm}>{\centering\arraybackslash}m{1.38cm}|>{\centering\arraybackslash}m{0.75cm}>{\centering\arraybackslash}m{0.75cm}>{\centering\arraybackslash}m{0.9cm}>{\centering\arraybackslash}m{0.93cm}|>{\centering\arraybackslash}m{0.75cm}>{\centering\arraybackslash}m{0.75cm}>{\centering\arraybackslash}m{0.9cm}>{\centering\arraybackslash}m{0.93cm}}
    \specialrule{.1em}{.05em}{0.0em}
        & &  \multicolumn{4}{c|}{4D-DRESS (cloth)}  & \multicolumn{4}{c}{4D-DRESS (cloth + human)} \\
        $L_{\text{cloth-SDS}}$ & $L_{\text{human-SDS}}$ & CD$^{\downarrow}$ & NC$^{\downarrow}$ & PSNR$^{\uparrow}$ & LPIPS$^{\downarrow}$ & CD$^{\downarrow}$ & NC$^{\downarrow}$ & PSNR$^{\uparrow}$ & LPIPS$^{\downarrow}$ \\
        \hline
        \xmark & \xmark & 4.704 & 0.041 & 28.151 & 0.047 & 3.635 & 0.086 & 22.895 & 0.083 \\
        {\fontsize{8}{16}\selectfont $\rightarrow L_{\text{SD-SDS}}$} & \cmark & 4.922 & 0.040 & 30.310 & 0.041 & 3.711 & 0.085 & 23.267 & 0.068 \\
        \cmark  & {\fontsize{8}{16}\selectfont $\rightarrow L_{\text{SD-SDS}}$} & 3.955 & \textbf{0.037} & 31.295 & 0.034 & 3.601 & 0.085 & 23.389 & 0.067 \\
        \cmark & \cmark & \textbf{3.902} & \textbf{0.037} & \textbf{31.582} & \textbf{0.033} & \textbf{3.292} & \textbf{0.079} & \textbf{23.921} & \textbf{0.065} \\
        \specialrule{.1em}{-0.05em}{-0.05em}
    \end{tabular}
    \vspace*{-1mm}
}
\end{center}
    \vspace*{-1.4em}
    \caption{
    \textbf{Ablation studies for cloth and human SDS loss.}
    }
    \vspace*{-0.1em}
    \label{tab:abla_sdsloss}
\end{table}
\begin{table}[t]
\def\arraystretch{1.48}
\renewcommand{\tabcolsep}{0.8mm}
\footnotesize
\begin{center}
\scalebox{0.77}{
    \begin{tabular}{>{\raggedright\arraybackslash}m{5.2cm}|>{\centering\arraybackslash}m{2.4cm}>{\centering\arraybackslash}m{2.4cm}}
    \specialrule{.1em}{.05em}{0.0em}
        & \multicolumn{2}{c}{4D-DRESS (cloth)}  \\
        Networks & CLIP-Score$^{\uparrow}$ & Cloth-IoU$^{\uparrow}$  \\
        \hline
        StableDiffusion~\cite{rombach2022high} & 0.696 & 0.112 \\
        HumanDiffusion~\cite{zhang2023adding} & 0.683 & 0.454  \\
        \textbf{ClothDiffusion (Ours)} & \textbf{0.713} & \textbf{0.548} \\
        \specialrule{.1em}{-0.05em}{-0.05em}
    \end{tabular}
    \vspace*{-1mm}
}
\end{center}
    \vspace*{-1.4em}
    \caption{
    \textbf{Ablation studies for ClothDiffusion network.}
    }
    \vspace*{-0.6em}
    \label{tab:abla_clothdiffusion}
\end{table}

\begin{figure*}[t!]
  \centering
  \includegraphics[width=1.0\linewidth]{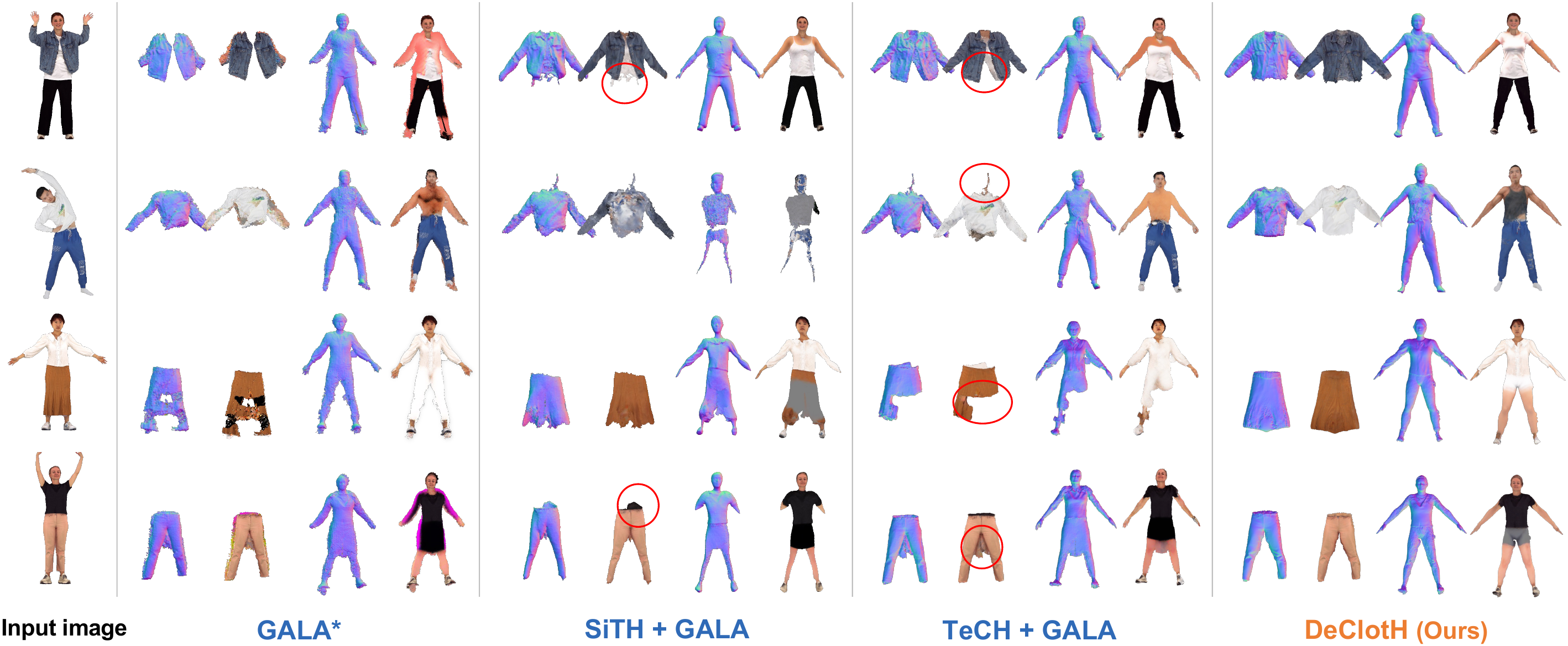}
  \vspace*{-1.6em}
  \caption{\textbf{Qualitative comparison with 3D cloth decomposition and 3D clothed human reconstruction methods: GALA$^{*}$~\cite{kim2024gala}, SiTH~\cite{ho2024sith}+GALA~\cite{kim2024gala}, and TeCH~\cite{huang2024tech}+GALA~\cite{kim2024gala}, on 4D-DRESS~\cite{wang20244d} and THuman2.0~\cite{tao2021function4d}.}
  $*$ denotes the algorithm is modified to take a single image as input instead of a 3D scan.
  We highlight their representative failure cases with red circles.
  }
   \vspace*{+0.2em}
  \label{fig:qual_decomposition}
\end{figure*}
\begin{table*}[t]
\def\arraystretch{1.6}
\renewcommand{\tabcolsep}{0.8mm}
\footnotesize
\begin{center}
\scalebox{0.755}{
    \begin{tabular}{>{\raggedright\arraybackslash}m{3.0cm}|>{\centering\arraybackslash}m{1.05cm}>{\centering\arraybackslash}m{1.05cm}>{\centering\arraybackslash}m{1.10cm}>{\centering\arraybackslash}m{1.10cm}|>{\centering\arraybackslash}m{1.05cm}>{\centering\arraybackslash}m{1.05cm}>{\centering\arraybackslash}m{1.10cm}>{\centering\arraybackslash}m{1.10cm}|>{\centering\arraybackslash}m{1.05cm}>{\centering\arraybackslash}m{1.05cm}>{\centering\arraybackslash}m{1.10cm}>{\centering\arraybackslash}m{1.10cm}|>{\centering\arraybackslash}m{1.05cm}>{\centering\arraybackslash}m{1.05cm}>{\centering\arraybackslash}m{1.10cm}>{\centering\arraybackslash}m{1.10cm}}
    \specialrule{.1em}{.05em}{0.0em}
        &  \multicolumn{4}{c|}{4D-DRESS (cloth)} & \multicolumn{4}{c|}{4D-DRESS (cloth + human)} & \multicolumn{4}{c|}{THuman2.0 (cloth)} & \multicolumn{4}{c}{THuman2.0 (cloth + human)} \\
        Methods  & CD$^{\downarrow}$ & NC$^{\downarrow}$ & PSNR$^{\uparrow}$ & LPIPS$^{\downarrow}$ & CD$^{\downarrow}$ & NC$^{\downarrow}$ & PSNR$^{\uparrow}$ & LPIPS$^{\downarrow}$ & CD$^{\downarrow}$ & NC$^{\downarrow}$ & PSNR$^{\uparrow}$ & LPIPS$^{\downarrow}$ & CD$^{\downarrow}$ & NC$^{\downarrow}$ & PSNR$^{\uparrow}$ & LPIPS$^{\downarrow}$ \\
        \hline
        GALA$^{*}$~\cite{kim2024gala} & 5.790 & 0.049 & 25.363 & 0.068 & 3.451 & 0.102 & 19.490 & 0.117 & 2.211 & 0.059 & 26.553 & 0.050 & 2.132 & 0.118 & 22.183 & 0.079 \\
        SiTH~\cite{ho2024sith} + GALA~\cite{kim2024gala} & 6.980 & 0.046 & 27.725 & 0.064 & 3.737 & 0.087 & 22.009 & 0.102 & 2.486 & 0.062 & 28.615 & 0.056 & 1.828 & 0.094 & 25.009 & 0.060 \\
        TeCH~\cite{huang2024tech} + GALA~\cite{kim2024gala} & 5.043 & 0.039 & 29.123 & 0.045 & 3.334 & 0.083 & 23.140 & 0.076 & 2.112 & 0.051 & 29.584 & 0.044 & 1.900 & 0.091 & \textbf{25.618} & 0.048 \\
        \textbf{\ourmethod~(Ours)} & \textbf{3.902} & \textbf{0.037} &   \textbf{31.582} & \textbf{0.033} & \textbf{3.292} & \textbf{0.079} & \textbf{23.921} & \textbf{0.065} & \textbf{1.756} & \textbf{0.044} & \textbf{30.612} & \textbf{0.032} & \textbf{1.812} & \textbf{0.089} & 25.421 & \textbf{0.046} \\
        \specialrule{.1em}{-0.05em}{-0.05em}
    \end{tabular}
}
\end{center}
    \vspace*{-1.4em}
    \caption{
        \textbf{Quantitative comparison with existing 3D cloth decomposition and 3D clothed human reconstruction methods.}
    }
    \vspace*{-0.3em}
    \label{tab:eval_decomp}
\end{table*}

\noindent\textbf{Ablation on ClothDiffusion network.}
\cref{tab:abla_clothdiffusion} shows our proposed ClothDiffusion is superior to other diffusion models in generating cloth images.
StableDiffusion~\cite{rombach2022high} and HumanDiffusion~\cite{zhang2023adding} commonly produce images that contain not only cloth but also other contents (\textit{e.g.}, human body and background scene).
Thus, StableDiffusion and HumanDiffusion have a low CLIP-Score~\cite{hessel2021clipscore} on text prompts that describe clothes only.
On the other hand, ClothDiffusion has a high CLIP-Score since it is specialized to generate cloth images without including other contents.
Additionally, ClothDiffusion achieves the highest Cloth-IoU among the diffusion models, since ClothDiffusion takes a cloth silhouette with a human skeleton as a condition, accurately guiding the region where the cloth will be generated.
Such superiority of ClothDiffusion in generating cloth images is beneficial in accurate guidance for reconstructing 3D cloth.

\subsection{Comparison with state-of-the-art methods}
We compare our method with recent 3D cloth decomposition and 3D clothed human reconstruction methods: GALA$^{\ast}$~\cite{kim2024gala}, SiTH~\cite{ho2024sith} + GALA~\cite{kim2024gala}, and TeCH~\cite{huang2024tech} + GALA~\cite{kim2024gala}.
GALA$^{\ast}$ is a modified version of the original GALA to take a single image as input instead of a 3D scan.
Specifically, GALA$^{\ast}$ only applies loss functions corresponding to the front view, ignoring other view directions.
SiTH + GALA and TeCH + GALA are two-stage reconstruction methods that first reconstruct a 3D clothed human, as one unified mesh, and decompose the 3D cloth mesh from the reconstructed 3D clothed human mesh.

\cref{fig:qual_decomposition} and \cref{tab:eval_decomp} show the superior performance of our \ourmethod~compared to the prior arts on 4D-DRESS~\cite{wang20244d} and THuman2.0~\cite{tao2021function4d}. 
GALA$^{\ast}$~\cite{kim2024gala} highly relies on the front view, which results in undesirable appearances, especially when the cloth is occluded.
On the other hand, \ourmethod~produces much more robust results from occlusion by using 3D template models of cloth and human body as strong geometric priors for reconstruction.
The two-stage reconstruction methods, SiTH~\cite{ho2024sith} + GALA~\cite{kim2024gala} and TeCH~\cite{huang2024tech} + GALA~\cite{kim2024gala}, also suffer from undesirable artifacts of reconstruction results, such as torn clothes.
The reason for such artifacts is that the 3D geometric error of the first stage, 3D clothed human reconstruction, fatally propagates to the second stage, 3D cloth decomposition.
The 3D clothed human reconstruction contains inevitable 3D geometric errors because of the ill-posedness of reconstruction.
The error of the first stage would be a bad source for the 3D cloth decomposition method, resulting in a failure of decomposition.
Compared to the two-stage reconstruction methods, \ourmethod~is a one-stage reconstruction method that is free from the above issue.
Additionally, while the two-stage methods do not consider the geometries of the cloth and human body during the reconstruction, \ourmethod~effectively guides the geometries by leveraging 3D template models and ClothDiffusion.
We provide further discussion in the supplementary material.
\vspace*{+0.65em}
\section{Conclusion}
We introduce~\ourmethod, a novel and powerful framework that reconstructs decomposable 3D cloth and human body from a single image.
Based on 3D template models of cloth and human body, our proposed template regularization loss and cloth diffusion model effectively infer geometry and texture in the invisible regions of 3D cloth and human body.
As a result, our framework significantly outperforms baseline methods, qualitatively and quantitatively.

\vspace*{+0.7em}
\noindent\textbf{Acknowledgements.}
This work was supported in part by the IITP grants [No.2021-0-01343, Artificial Intelligence Graduate School Program (Seoul National University), No. 2021-0-02068, and  No.2023-0-00156], and the NOTIE grant (No. RS-2024-00432410) by the Korean government.

\clearpage
\setcounter{page}{1}
\maketitlesupplementary
\vspace*{+0.5em}

\setcounter{section}{0}
\setcounter{table}{0}
\setcounter{figure}{0}
\renewcommand{\thesection}{S\arabic{section}}   
\renewcommand{\thetable}{S\arabic{table}}   
\renewcommand{\thefigure}{S\arabic{figure}}

In this supplementary material, we present additional technical details and more experimental results that could not be included in the main manuscript due to the lack of pages.
The contents are summarized below:
\begin{itemize}
\vspace{2.0mm}
\item \ref{sec:suppl_control}. Controlling reconstruction results
\item \ref{sec:suppl_deformation}. Evaluation of pose deformation
\item \ref{sec:suppl_por}. Evaluation with POR Score
\item \ref{sec:suppl_imple_details}. Implementation details

\item \ref{sec:suppl_analysis}. Discussion of two-stage reconstruction 
\item \ref{sec:suppl_limitation}. Limitations and future works
\item \ref{sec:suppl_more_qual}. More qualitative results
\end{itemize}

\section{Controlling reconstruction results}
\label{sec:suppl_control}
\begin{figure}[t!]
  \vspace*{+1.8em}
  \centering
  \includegraphics[width=1.0\linewidth]{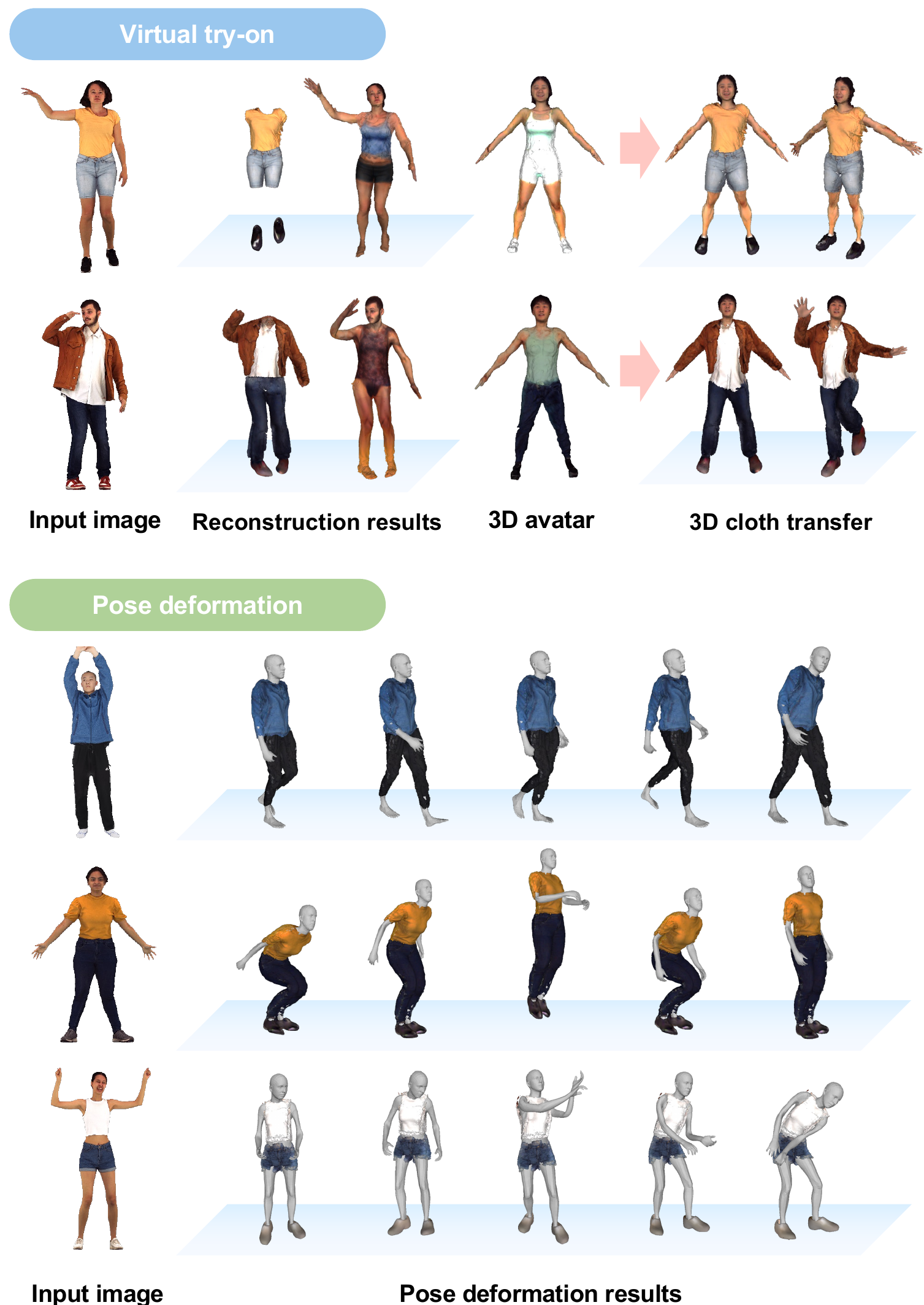}
  \vspace*{-1.2em}
  \caption{\textbf{Examples for controlling 3D reconstruction results.}
  Our reconstruction results are editable, such as virtual try-on and pose deformation.
  }
  \vspace*{-0.6em}
  \label{fig:suppl_controlling}
\end{figure}

Our proposed \ourmethod~has the advantage of easily modifying the reconstructed results for virtual try-on and pose deformation.
\cref{fig:suppl_controlling} illustrates the examples of controlling the reconstruction results.
First, we can transfer the reconstructed 3D clothes into a new 3D avatar, by fitting 3D clothes based on the SMPL+H human model (blue part of the figure).
Second, by forwarding new SMPL+H pose parameters to the linear blend skinning (LBS) of our pipeline, we can animate the reconstruction results (green part of the figure).
Like these examples, reconstructing separate 3D geometries is highly useful for applying human reconstruction systems for various downstream applications.

\section{Evaluation of pose deformation}
\label{sec:suppl_deformation}
We demonstrate that our \ourmethod~is also superior to existing methods in applying pose deformation to reconstruction results.
For the evaluation, we deform the reconstruction results with GT human pose parameters of 4D-DRESS~\cite{wang20244d} test set.
4D-DRESS contains sequences of 3D cloth and human scans, driven by human pose parameters.
Using these pose parameters, we deform the reconstructed meshes to follow the first pose of each sequence.
Then, we evaluate the deformed meshes based on the GT 3D scans corresponding to the first pose.
The evaluation results are shown in \cref{tab:suppl_pose_deformation}, indicating our \ourmethod~has the advantage over other methods for animating reconstruction results with novel human poses.

\section{Evaluation with POR Score}
\label{sec:suppl_por}
We provide more quantitative comparison results through POR Score (pixel-wise object removal score) proposed by Kim~\etal~\cite{kim2024gala}.
The POR Score is devised to evaluate the quality of 3D decomposition in the absence of 3D cloth GT scans.
This metric measures the proportion of remaining cloth pixels in rendered human body images, after performing cloth decomposition.
Specifically, given a reconstructed 3D human body with the target cloth removed, we render 30 images using uniformly distributed camera viewpoints.
Subsequently, we run the off-the-shelf image segmentation method, SAM~\cite{kirillov2023segment}, to obtain the cloth segmentation corresponding to the cloth prompt.
Here, the cloth prompts are acquired by running the image captioning method, BLIP~\cite{li2022blip}.
From the obtained segmentations, the POR Score measures the ratio of pixels classified as the target cloth in the image.
A lower POR Score indicates better performance of the 3D cloth decomposition.
As shown in \cref{tab:suppl_por_score}, our framework also outperforms the other methods in POR Score, which demonstrates that \ourmethod~achieves better results in 3D cloth and human body decomposition.

\begin{table}[t]
\def\arraystretch{1.48}
\renewcommand{\tabcolsep}{0.85mm}
\footnotesize
\begin{center}
\scalebox{0.76}{
    \begin{tabular}{>{\raggedright\arraybackslash}m{2.9cm}|>{\centering\arraybackslash}m{0.8cm}>{\centering\arraybackslash}m{0.8cm}>{\centering\arraybackslash}m{0.85cm}>{\centering\arraybackslash}m{0.85cm}|>{\centering\arraybackslash}m{0.8cm}>{\centering\arraybackslash}m{0.8cm}>{\centering\arraybackslash}m{0.85cm}>{\centering\arraybackslash}m{0.85cm}}
    \specialrule{.1em}{.05em}{0.0em}
        & \multicolumn{4}{c|}{4D-DRESS (cloth)}  & \multicolumn{4}{c}{4D-DRESS (cloth + human)} \\
        Methods & CD$^{\downarrow}$ & NC$^{\downarrow}$ & PSNR$^{\uparrow}$ & LPIPS$^{\downarrow}$ & CD$^{\downarrow}$ & NC$^{\downarrow}$ & PSNR$^{\uparrow}$ & LPIPS$^{\downarrow}$ \\
        \hline
        GALA$^{*}$~\cite{kim2024gala}  & 5.251 & 0.044 & 25.390 & 0.069 & 2.844 & 0.088 & 19.454 & 0.117 \\
        SiTH~\cite{ho2024sith} + GALA~\cite{kim2024gala} & 6.560 & 0.042 & 27.578 & 0.065 & 3.364 & 0.078 & 21.886 & 0.102 \\
        TeCH~\cite{huang2024tech} + GALA~\cite{kim2024gala} & 4.425 & 0.033 & 29.271 & 0.044 & 2.422 & 0.059 & 23.276 & 0.070 \\
        \textbf{\ourmethod~(Ours)} & \textbf{2.782} & \textbf{0.030} & \textbf{31.489} & \textbf{0.033} & \textbf{2.271} & \textbf{0.055} & \textbf{23.369} & \textbf{0.067} \\
        \specialrule{.1em}{-0.05em}{-0.05em}
    \end{tabular}
}
\end{center}
    \vspace*{-1.5em}
    \caption{
    \textbf{Quantitative comparisons of pose deformation with 3D cloth decomposition and 3D cloth human reconstruction methods.}
    }
    \vspace*{-0.1em}
    \label{tab:suppl_pose_deformation}
\end{table}
\begin{table}[t]
\def\arraystretch{1.56}
\renewcommand{\tabcolsep}{0.8mm}
\footnotesize
\begin{center}
\scalebox{0.8}{
    \begin{tabular}{>{\raggedright\arraybackslash}m{4.1cm}|>{\centering\arraybackslash}m{2.6cm}}
    \specialrule{.1em}{.05em}{0.0em}
        Methods & POR Score$^{\downarrow}$ \\
        \hline
        GALA$^{*}$~\cite{kim2024gala} & 0.418 \\
         SiTH~\cite{ho2024sith} + GALA~\cite{kim2024gala} & 0.246 \\
        TeCH~\cite{huang2024tech} + GALA~\cite{kim2024gala} & 0.225 \\
        \textbf{\ourmethod~(Ours)} & \textbf{0.218} \\
        \specialrule{.1em}{-0.05em}{-0.05em}
    \end{tabular}
    \vspace*{-1mm}
}
\end{center}
    \vspace*{-1.4em}
    \caption{
    \textbf{Quantitative comparisons of POR Score~\cite{kim2024gala}  with 3D cloth decomposition and 3D cloth human reconstruction methods, on 4D-DRESS~\cite{wang20244d}.}
    }
    \vspace*{-0.6em}
    \label{tab:suppl_por_score}
\end{table}

\section{Implementation details}
\label{sec:suppl_imple_details} 
\noindent\textbf{Network architecture.}
The DMTets, which are optimized at the geometry stage, are implemented by using two fully-connected layers with 32 hidden dimensions and ReLU activations.
The DMTets take the 3D vertex coordinates of the tetrahedral grid ($\mathbf{X}_T$, $T$) as input, where the coordinates are normalized between -0.5 and 0.5.
Then, the coordinates are encoded by a hash positional encoding~\cite{muller2022instant} with a maximum resolution of 1024 and 16 resolution levels.
The MLP networks, which are optimized at the texture stage, are implemented by using a fully-connected layer with 32 hidden dimension and ReLU activations.
The MLP networks take the mesh coordinates as input, after applying the hash positional encoding with a maximum resolution of 2048.
Additionally, we implement a MLP network, which takes camera parameter $\mathbf{k}$ and produces adaptive background colors of the rendering pipeline, using two fully-connected layers.

\noindent\textbf{Optimization details.}
PyTorch~\cite{paszke2017automatic} is used for the implementation. 
In both the geometry and texture stages, we use Adam optimizer~\cite{kingma2014adam} with 4000 optimization steps.
The initial learning rate is set to $0.001$ and reduced by an exponential scheduler, $\eta = 0.001 \times 0.1^{step/4000}$.
During the optimization process, we render 3D cloth and human body based on the spherical coordinate system, ($r$, $\theta$, $\phi$), where $r$ denotes the distance from the spherical origin, $\theta$ denotes the elevation angle, and $\phi$ denotes the azimuth angle. 
We set $r \in [0.7, 1.3]$, $\theta \in [-30^{\circ}, 30^{\circ}]$, and $\phi \in [-180^{\circ}, 180^{\circ}]$, with uniform sampling.
To capture fine details of human faces, we additionally use zoomed-in camera views for the rendering.
Specifically, we set the spherical origin to the 3D position of SMPL+H head keypoint, $r \in [0.3, 0.4]$, $\theta \in [-90^{\circ}, 90^{\circ}]$, and $\phi \in [-90^{\circ}, 90^{\circ}]$.
All the experiments are conducted with an NVIDIA Quadro RTX 8000 GPU.

\noindent\textbf{Training details for ClothDiffusion.}
To train ClothDiffusion described in \cref{sec:clothdiffusion}, we adopt StableDiffusion~\cite{rombach2022high} in version 1.5.
The weights of ClothDiffusion are updated by Adam optimizer~\cite{kingma2014adam} with 200k training steps and a mini-batch size of 8.
The learning rate is set to $10^{-5}$. 
We train the model with an NVIDIA Quadro RTX 8000 GPU.

\begin{figure}[t!]
  \centering
  \includegraphics[width=1.0\linewidth]{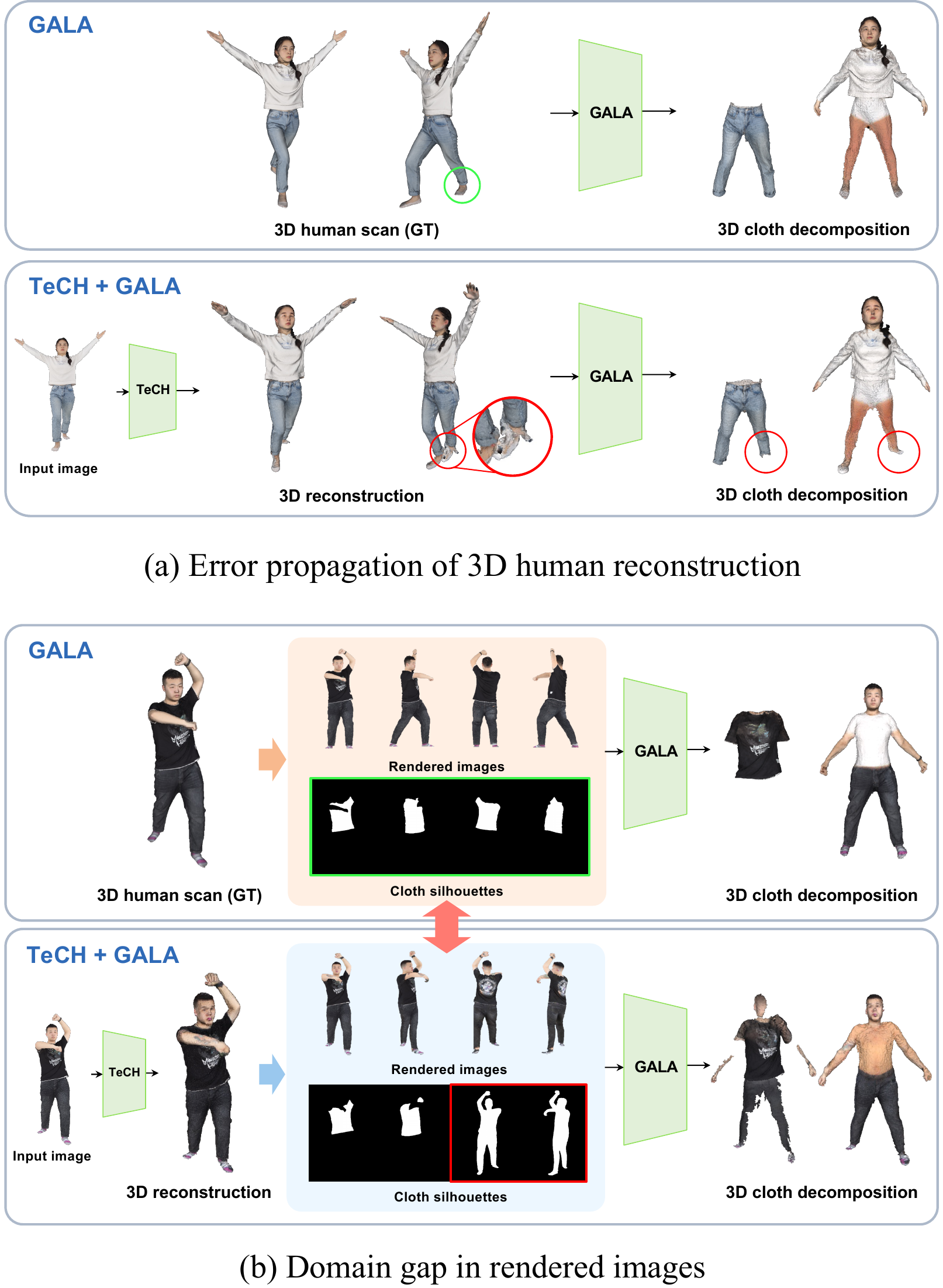}
  \vspace*{-1.3em}
  \caption{\textbf{Failure examples of two-stage reconstruction methods}: (a) propagation of 3D reconstruction error and (b) domain gap in rendered images.
  }
  \vspace*{-0.5em}
  \label{fig:discussion_two_stage}
\end{figure}

\section{Discussion of two-stage reconstruction}
\label{sec:suppl_analysis}
In this section, we provide a deep discussion about the advantages of our \ourmethod~compared to the two-stage reconstruction methods, SiTH~\cite{ho2024sith} + GALA~\cite{kim2024gala} and TeCH~\cite{huang2024tech} + GALA~\cite{kim2024gala}.
We suggest that the two-stage reconstruction methods have two drawbacks: 1) propagation of 3D reconstruction error and 2) domain gap in the rendered images.

\noindent\textbf{Error propagation of 3D human reconstruction.}
\cref{fig:discussion_two_stage} (a) illustrates that the 3D geometric errors from the 3D human reconstruction significantly affect 3D cloth decomposition errors.
In the first row of the figure, GALA~\cite{kim2024gala} accurately decomposes 3D cloth when provided with a 3D human GT scan, which is naturally free of geometric artifacts.
On the other hand, in the second row of the figure, TeCH~\cite{huang2024tech} produces the 3D geometric error in reconstructing the ankle part, leading to the annihilation of ankle parts in the 3D cloth decomposition.
The primary discrepancy lies in 3D human reconstruction methods overlooking the geometric relationship between the cloth and the human body, leading to overly thick or thin reconstructions.
While these thick or thin reconstructions appear visually acceptable, they are critically detrimental to 3D cloth decomposition. 
Unlike the two-stage approach, our \ourmethod~considers the volumetric space for 3D cloth decomposition during the reconstruction process. 
Therefore, \ourmethod~is free from error propagation issue and provides accurate reconstructions of 3D cloth and human body.

\noindent\textbf{Domain gap in rendered image.}
\cref{fig:discussion_two_stage} (b) shows that there is a domain gap issue in rendered images between real-world and reconstructed 3D avatars, leading to wrong 3D cloth decomposition.
The 3D cloth decomposition method, GALA~\cite{kim2024gala}, runs based on the cloth silhouettes from the rendered images of a given 3D avatar.
Here, the cloth silhouettes are acquired through the image segmentation method, SAM~\cite{kirillov2023segment}.
GALA (first row of the figure) results in the accurate decomposition result by utilizing correct cloth silhouettes for all rendered images.
In contrast, TeCH+GALA (second row of the figure) produces the erroneous result since the cloth segmentation often fails.
We conjecture that the failure of the cloth segmentation is the domain gap in rendered images.
Based on the 3D human reconstruction results of TeCH~\cite{huang2024tech}, its rendered images have artificial appearances compared to real images.
Such artificial appearances adversely affect the decomposition of 3D clothes from the 3D human reconstruction results.
On the other hand, our proposed \ourmethod~is a one-stage method that does not require performing segmentation for rendered images.
Therefore, our \ourmethod~does not have the domain gap issue, which is an advantage over the two-stage reconstruction methods.

\section{Limitations and future works}
\label{sec:suppl_limitation}
\begin{figure}[t!]
  \centering
  \includegraphics[width=0.9\linewidth]{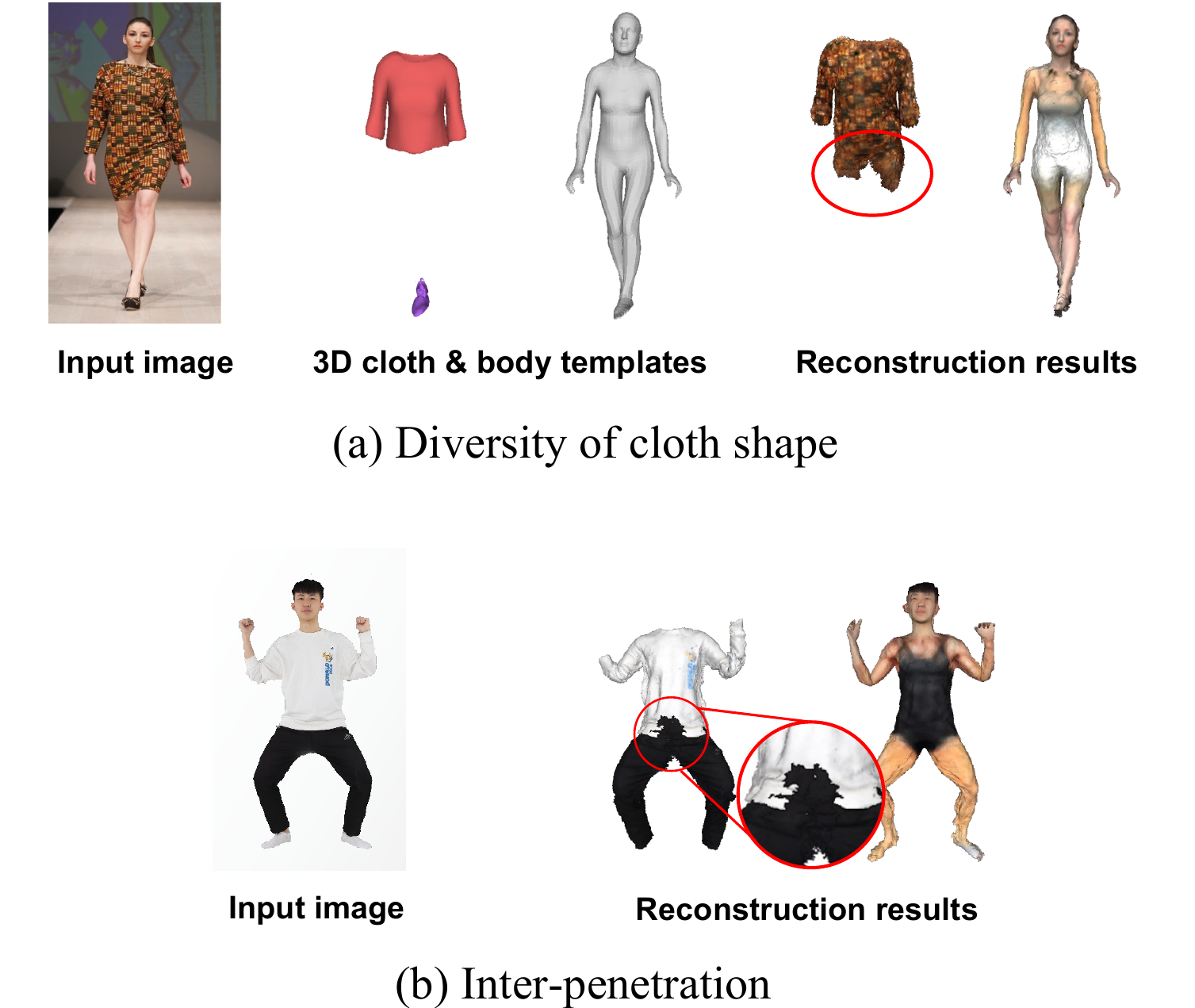}
  \vspace*{-0.0em}
  \caption{\textbf{Failure cases of our proposed framework.}
  }
  \vspace*{-0.3em}
  \label{fig:suppl_limitations}
\end{figure}

\noindent\textbf{Diversity of cloth shape.}
There is a limitation in reconstructing diverse cloth types (\textit{e.g.}, dress), as shown in \cref{fig:suppl_limitations} (a). 
This is mainly due to the expression power of the cloth template model (\textit{i.e.}, SMPLicit~\cite{corona2021smplicit}).
Most of the existing cloth template models~\cite{bhatnagar2019multi,jiang2020bcnet,moon20223d,de2023drapenet,corona2021smplicit} have difficulty in modeling the wide variety of 3D cloth geometries in the real world.
Thereby, for several uncommon clothes, predicting 3D cloth templates often fails, and \ourmethod's reconstruction based on the cloth templates also produces erroneous results.
Improving the expression power of cloth template models should be a future research direction.

\noindent\textbf{Inter-penetration.}
\cref{fig:suppl_limitations} (b) shows that our framework often suffers from inter-penetration in reconstructed 3D clothes.
This inter-penetration issue is extremely challenging, as it requires reasoning not only about the geometric relationship between the cloth and the human body, but also among different clothes.
Accordingly, we aim to extend our framework to efficiently reconstruct 3D clothes while overcoming the inter-penetration issue.

\section{More qualitative results}
\label{sec:suppl_more_qual}
\begin{figure*}[t!]
  \centering
  \includegraphics[width=0.97\linewidth]{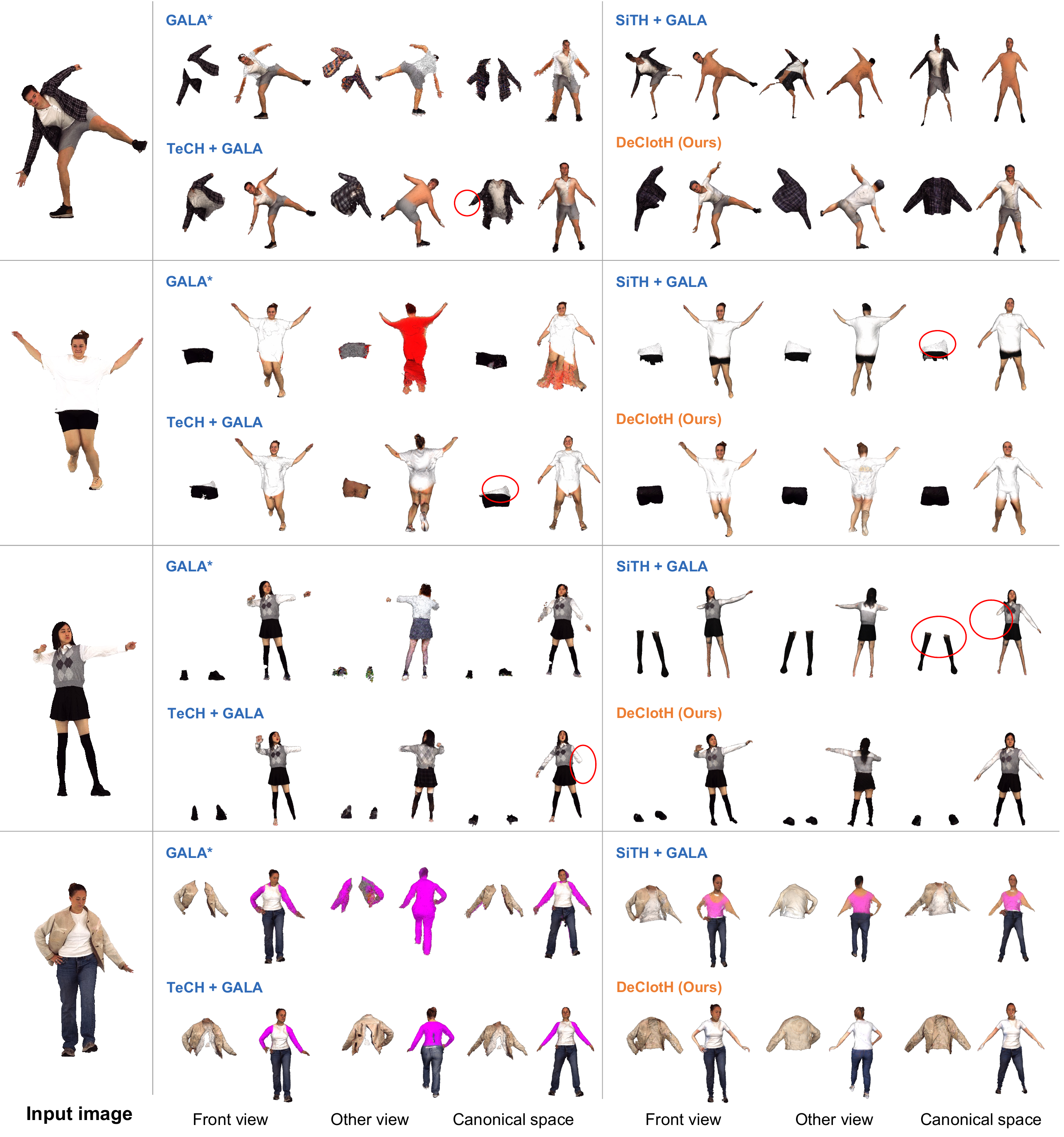}
  \vspace*{-0.2em}
  \caption{\textbf{Qualitative comparison with 3D cloth decomposition and 3D clothed human reconstruction methods: GALA$^{*}$~\cite{kim2024gala}, SiTH~\cite{ho2024sith}+GALA~\cite{kim2024gala}, and TeCH~\cite{huang2024tech}+GALA~\cite{kim2024gala}, on 4D-DRESS~\cite{wang20244d}.}
  $*$ denotes the algorithm is modified to take a single image as input instead of a 3D scan.
  We highlight their representative failure cases with red circles.
  }
  \label{fig:qual_decomposition_1}
\end{figure*}

\begin{figure*}[t!]
  \centering
  \includegraphics[width=0.97\linewidth]{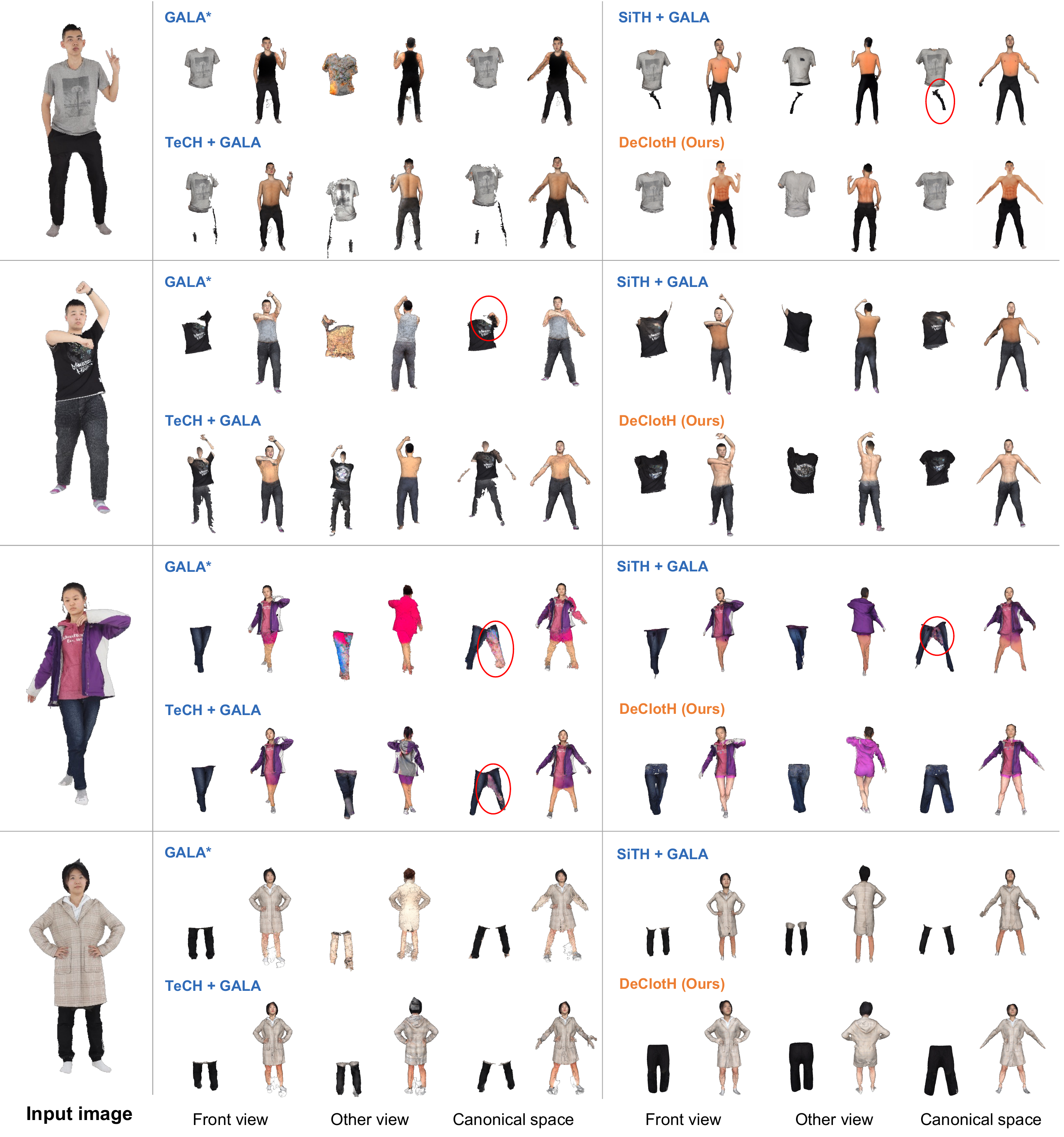}
  \vspace*{-0.2em}
  \caption{\textbf{Qualitative comparison with 3D cloth decomposition and 3D clothed human reconstruction methods: GALA$^{*}$~\cite{kim2024gala}, SiTH~\cite{ho2024sith}+GALA~\cite{kim2024gala}, and TeCH~\cite{huang2024tech}+GALA~\cite{kim2024gala}, on THuman2.0~\cite{tao2021function4d}.}
  $*$ denotes the algorithm is modified to take a single image as input instead of a 3D scan.
  We highlight their representative failure cases with red circles.
  }
  \label{fig:qual_decomposition_2}
\end{figure*}

\begin{figure*}[t!]
  \centering
  \includegraphics[width=0.97\linewidth]{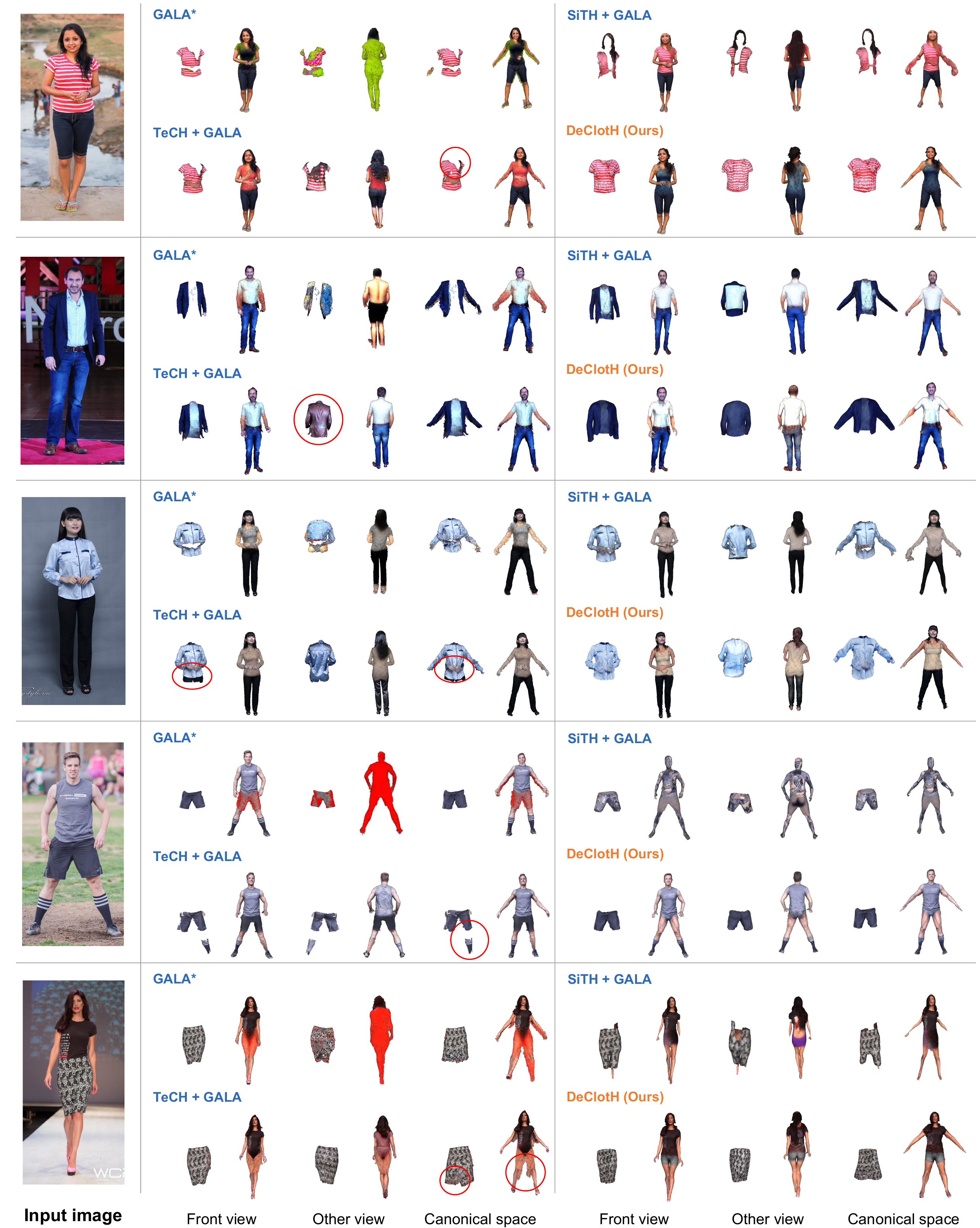}
  \vspace*{-0.2em}
  \caption{\textbf{Qualitative comparison with 3D cloth decomposition and 3D clothed human reconstruction methods: GALA$^{*}$~\cite{kim2024gala}, SiTH~\cite{ho2024sith}+GALA~\cite{kim2024gala}, and TeCH~\cite{huang2024tech}+GALA~\cite{kim2024gala}, on in-the-wild images.}
  $*$ denotes the algorithm is modified to take a single image as input instead of a 3D scan.
  We highlight their representative failure cases with red circles.
  }
  \label{fig:qual_decomposition_3}
\end{figure*}

We provide more qualitative comparisons of 3D clothing reconstruction on 4D-DRESS~\cite{wang20244d} and THuman2.0~\cite{tao2021function4d}.
Figs.~\ref{fig:qual_decomposition_1} and~\ref{fig:qual_decomposition_2} show that our \ourmethod~produces far more accurate reconstructions of 3D cloth and human body compared to the prior arts. 
\cref{fig:qual_decomposition_3} demonstrates that \ourmethod~also achieves superior reconstruction performance on in-the-wild images.

\begin{figure*}[t!]
  \centering
  \includegraphics[width=1.0\linewidth]{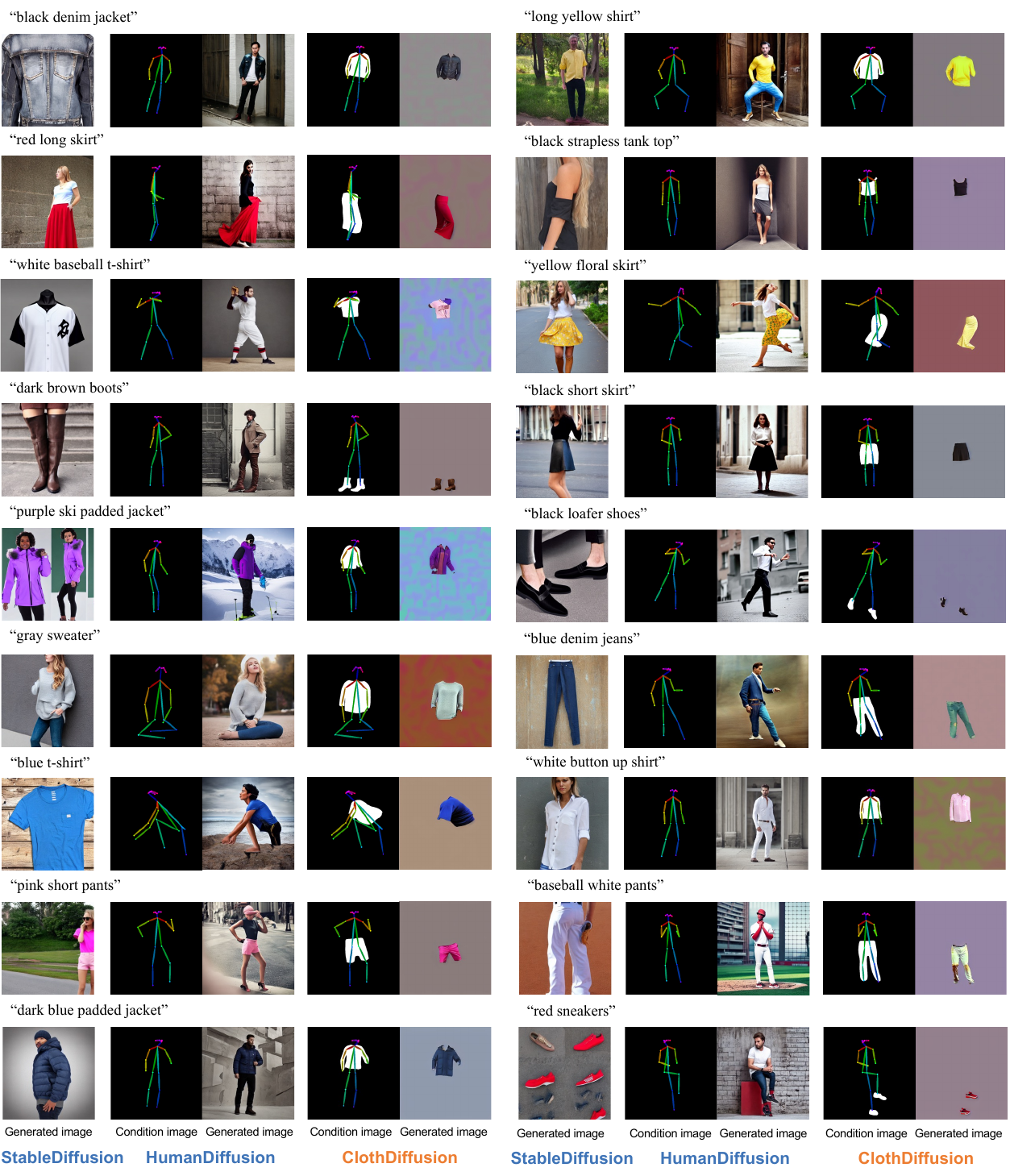}
  \vspace*{-1.6em}
  \caption{\textbf{Qualitative comparison of cloth image generation between StableDiffusion~\cite{rombach2022high}, HumanDiffusion~\cite{zhang2023adding}, and our proposed ClothDiffusion.}
  }
  \label{fig:qual_clothdiffusion}
\end{figure*}

\cref{fig:qual_clothdiffusion} shows the qualitative comparison of StableDiffusion~\cite{rombach2022high}, HumanDiffusion~\cite{zhang2023adding}, and our proposed ClothDiffusion.
Compared to StableDiffusion and HumanDiffusion, ClothDiffusion specializes in cloth image generation, excluding other contents.
Additionally, ClothDiffusion accurately generates cloth images in the desired regions corresponding to the condition images.

\clearpage
\clearpage
{
    \small
    \bibliographystyle{ieeenat_fullname}
    \bibliography{egbib}
}
\end{document}